\documentclass[12pt]{article}

\usepackage[a4paper,top=3cm,bottom=2cm,left=2cm,right=2cm,marginparwidth=1.25cm]{geometry}
\usepackage{subfigure}
\usepackage[utf8]{inputenc} 
\usepackage[T1]{fontenc}    
\usepackage{url}            
\usepackage{booktabs}       
\usepackage{amsfonts}       
\usepackage{nicefrac}       
\usepackage{microtype}      
\newcommand {\myvec}[1] {{\mbox{\boldmath $#1$}}}

\usepackage{amssymb}
\usepackage{amsmath}
\usepackage{mathtools}
\usepackage{amsthm}
\usepackage{color}
\usepackage{xr}
\usepackage{adjustbox}

\newcommand{\beginsupplement}{%
        \setcounter{table}{0}
        \renewcommand{\thetable}{S\arabic{table}}%
        \setcounter{figure}{0}
        \renewcommand{\thefigure}{S\arabic{figure}}         \setcounter{section}{0}
        \renewcommand{\thesection}{S\arabic{section}}
}
\makeatletter

\makeatletter
\newcommand{\distas}[1]{\mathbin{\overset{#1}{\kern\z@\sim}}}%
\newsavebox{\mybox}\newsavebox{\mysim}
\newcommand{\distras}[1]{%
  \savebox{\mybox}{\hbox{\kern3pt$\scriptstyle#1$\kern3pt}}%
  \savebox{\mysim}{\hbox{$\sim$}}%
  \mathbin{\overset{#1}{\kern\z@\resizebox{\wd\mybox}{\ht\mysim}{$\sim$}}}%
}
\theoremstyle{plain}

\usepackage{amscd}     
\usepackage{amsmath}   
\usepackage{amssymb}   
\usepackage{amsthm}    
\usepackage{comment}
\usepackage{adjustbox}

\usepackage{mathtools}

\usepackage{booktabs}

\newtheorem{Theorem}{Theorem}[section]

\theoremstyle{remark}

\newcommand{\eps}{\varepsilon}

\newcommand{\RR}{\mathbb{R}}

\title{Probabilistic Robust Autoencoders for Outlier Detection}

\author{ Ofir Lindenbaum $^{1 \ast }$ \and Yariv Aizenbud $^{2 \ast }$ \and Yuval Kluger$^{1 \dagger}$\\
\normalsize{$^{1}$ Bar-Ilan University, Israel;}\quad \normalsize{$^{2}$Yale University, USA;}\\
\normalsize{$^\dagger$Corresponding author. E-mail: yuval.kluger@yale.edu}\\\normalsize{Address: 333 Cedar St, New Haven, CT 06510, USA}\\
\normalsize{$^\ast$ These authors contributed equally.}
}

\date{}

\begin{document}

\maketitle
\begin{abstract}

Anomalies (or outliers) are prevalent in real-world empirical observations and potentially mask important underlying structures. 
Accurate identification of anomalous samples is crucial for the success of downstream data analysis tasks. To automatically identify anomalies, we propose Probabilistic Robust AutoEncoder (PRAE). 
PRAE aims to simultaneously remove outliers and identify a low-dimensional representation for the inlier samples. 
We first present the Robust AutoEncoder (RAE) objective as a minimization problem for splitting the data into inliers and outliers. Our objective is designed to exclude outliers while including a subset of samples (inliers) that can be effectively reconstructed using an AutoEncoder (AE). RAE minimizes the autoencoder's reconstruction error while incorporating as many samples as possible. 
This could be formulated via regularization by subtracting an $\ell_0$ norm counting the number of selected samples from the reconstruction term. Unfortunately, this leads to an intractable combinatorial problem. Therefore, we propose two probabilistic relaxations of RAE, which are differentiable and alleviate the need for a combinatorial search. We prove that the solution to the PRAE problem is equivalent to the solution of RAE. We use synthetic data to show that PRAE can accurately remove outliers in a wide range of contamination levels. Finally, we demonstrate that using PRAE for anomaly detection leads to state-of-the-art results on various benchmark datasets.
\end{abstract}

Unsupervised anomaly detection is a fundamental problem in data mining and machine learning. The goal is to identify unusual measurements in unlabeled datasets. 
Identifying anomalous samples is essential for empirical science in various fields, such as biology \cite{anom-bio}, geophysics \cite{bregman2021array}, engineering, and cyber-security \cite{anom-cyber}. 
Anomalies and outliers are samples that significantly deviate from the ``normal'' (often majority) observations.
A critical challenge is defining such normality and automatically identifying all abnormal measurements.

One solution for anomaly detection relies on the data's density. By estimating the data density, anomalies could be identified as samples drawn from the low probability density regions \cite{bishop1994novelty}. 
Density-based models include Local Outlier Factor (LOF) \cite{breunig2000lof}, or some of its variants \cite{jin2001mining,tang2002enhancing,jin2006ranking}. 
More recent probabilistic approaches include \cite{kriegel2009loop,constantinou2018pynomaly}. 
Other schemes \cite{aizenbud2015pca,ramaswamy2000efficient,angiulli2002fast,ghoting2008fast} rely on distances between samples to identify anomalies; the basic assumption is that normal points have dense neighborhoods while outliers are far from their neighbors. Another paradigm for anomaly detection is one-class-classification, where anomalies are identified as samples that significantly deviate from the major part of the data (i.e. the "one-class") \cite{chen2001one,ruff2018deep,perera2021one,deng2022anomaly}.   

\everypar{\looseness=-1}
High dimensional measurements may often be described by a low dimensional subspace or manifold~\cite{aizenbud2021non,roweis2000LLE,peterfreund2020local}.
By assuming that the normal samples lie near a low dimensional latent manifold, while outliers are diverse and do not follow the same manifold structure, the anomalies could be detected via a dimensionality reduction method, such as Principal Component Analysis (PCA) \cite{Pearson1901PCA}, or deep Autoencoders (AE) \cite{rumelhart1985learning,japkowicz1995novelty,lecun1989generalization}. 
Robust PCA schemes \cite{lerman2018overview} seek for a low dimensional linear subspace that ``best" fits the inliers. 
These models can identify anomalies and learn a reduced sup-space simultaneously; however, they are restricted to linear transformations. 
To overcome this limitation, several authors have proposed to use AEs \cite{chen2017outlier,zhou2017anomaly} to learn a valuable nonlinear mapping for detecting outliers. 

Generative models are powerful tools for learning the data distribution and, therefore, turn valid for anomaly detection \cite{zong2018deep,liu2019generative,du2019implicit,eduardo2020robust}. In the more relaxed semi-supervised setting, several authors \cite{hendrycks2018deep,wang2019effective,goyal2020drocc,reiss2021panda,hojjati2021dasvdd} have used 
deep neural networks to model the normal part of the data; and identify outliers as samples that deviate significantly from normal samples. Recently, there has been growing interest in anomaly detection in vision, self-supervision \cite{hendrycks2019using,bergman2020classification}, and transfer learning \cite{deecke2021transfer} can be used to learn feature maps that capture the normal part of the data.

In this work, we focus on unsupervised anomaly detection for general data (not necessarily images) and propose a novel Probabilistic Robust autoencoder (PRAE). We note that in this work we treat anomalies and outliers in the same way since we only assume that they deviate from the ``normal'' samples. PRAE can simultaneously remove outliers and learn a low-dimensional representation of the inlier samples. 
Our contributions are four folds: (1) We formulate the robust autoencoding (robust-AE) problem by incorporating an $\ell_0$ term penalizing the number of observations included in the AE's reconstruction loss. (2) We propose two probabilistic relaxations for robust AE and demonstrate that they could be effectively trained using standard optimization tools such as stochastic gradient descent. (3)
We show theoretically that the solution of the probabilistic relaxation is equivalent to the solution to the robust-AE problem. (4) We demonstrate using extensive simulations that PRAE outperforms leading anomaly detection methods in multiple settings.   
\section{Background}
{\bf Notation:} Throughout the paper we denote vectors using {{bold}} lowercase letters such as $\myvec{x}$. Scalars are denoted by lower case letters such as $y$. The $n^{th}$ vector-valued observation is denoted as
$\myvec{x}_n$ while $x[d]$ represents the $d^{th}$ feature (or entry) of the vector $\myvec{x}$. Matrices are denoted by {{bold}} uppercase letters $\myvec{X}$. The
$\ell_p$ norm of $\myvec{x}$ is denoted by $\|\myvec{x}\|_p$. 
\subsection{Autoencoder (AE)}
The AE is a multilayer neural network designed for dimensionality reduction. 
It comprises an encoder and decoder, which are typically symmetric and are trained jointly to minimize the reconstruction error of the data. The number of neurons in the latent space (the hidden layer between the encoder and decoder) controls the dimension of the reduced representation of the data. 
It has been shown that the reduced representation of a linear AE spans the same subspace as the principal components of the data \cite{plaut2018principal}. Although is has been suggested to use an AE without dimensionality reduction \cite{yong2022autoencoders}, in this work all AE have low dimensional latent space.

Given samples $X=\{\myvec{x}_1,...,\myvec{x}_N\},$ where $\myvec{x}_i \in \mathbb{R}^D$, the AE learns the reduced representation by minimizing the following reconstruction loss
\begin{eqnarray}
\label{eq:recon_loss}
\frac{1}{N} \sum_{i} \Big\| \myvec{x}_i - \widehat{\myvec{x}}_i\Big\|_2^2,
\end{eqnarray}
where the reconstructed vector $\widehat{\myvec{x}}_i$ is obtained as the output of the encoder $\myvec{\rho()}$ and decoder $\myvec{\psi()}$, i.e. $\widehat{\myvec{x}}_i=\myvec{\psi}(\myvec{\rho}(\myvec{x}_i))$.
The encoder-decoder pair are defined using a multi-layer neural network; this can be described using the following equations: 
\begin{eqnarray*}
 &{\myvec{\rho}}_\ell (\myvec{x})=\myvec{\sigma}\left(\myvec{W}^{\rho}_{\ell-1} {\myvec{\rho}}_{\ell-1}(\myvec{x})+\myvec{b}^{\rho}_{\ell-1}\right), \mbox{~~and~~}\\ &{\myvec{\psi}}_{\ell}(\myvec{z})=\myvec{\sigma}\left(\myvec{W}^{\psi}_{\ell-1} {\myvec{\psi}}_{\ell-1}(\myvec{z})+\myvec{b}^{\psi}_{\ell-1}\right),\  \ell=1,\dots,L, 
\end{eqnarray*}

where ${\myvec{\rho}}^{(0)}(\myvec{x})=\myvec{x}$ and ${\myvec{\psi}}_{0}(\myvec{z})=\myvec{z}=\myvec{\rho}_{L}(\myvec{x})$. The weights $\myvec{W}^{\rho}_{\ell}$, $\myvec{W}^{\psi}_{\ell}$ and biases $\myvec{b}^{\rho}_{\ell}$, $\myvec{b}^{\psi}_{\ell}$ at each layer $\ell$ are learnt by applying stochastic gradient descent to the reconstruction loss. 
The operator $\myvec{\sigma}$ is a nonlinear activation function applied in an element-wise fashion. 

\section{Method}
\subsection{Robust Autoencoder}\label{sec:sparse}
Given samples $\myvec{X}  = \{\myvec{x}_1,\ldots, \myvec{x}_N\}$, where $\myvec{x}_i \in \mathbb{R}^{D}$, we model the data by $\myvec{X}=\myvec{X}_{in} \cup \myvec{X}_{out}$, where $\myvec{X}_{in}$ are inliers and $\myvec{X}_{out}$ are outliers. 
We assume that $\myvec{X}_{in}$ can be approximated by some low dimensional structure. Our goal is to identify the inliers and outliers. 
We propose to use a regularized AE that simultaneously learns a low dimensional representation of the data and identifies the outliers.
We define an indicator vector $\myvec{b}\in \{0,1\}^N$ whose value $i$ indicates if the sample $\myvec{x}_i$ is an inlier ($b[i] = 1$) or an outlier ($b[i]=0$). 
To learn the parameters of the encoder-decoder pair ($\myvec{\rho()}$ and $\myvec{\psi()}$) while simultaneously identifying the inliers and outliers, we propose the following robust AE loss
\begin{equation}\label{eq:det_opt}
    L_d(\myvec{\psi}, \myvec{\rho}, \myvec{b}) = \sum_{i}  b[i] \Big\| \myvec{x}_i - \widehat{\myvec{x}}_i\Big\|_2^2 -\lambda \|\myvec{b}\|_0,
\end{equation}
where $\widehat{\myvec{x}}_i=\myvec{\psi}(\myvec{\rho}(\myvec{x}_i))$. 
The leading term in Eq. \eqref{eq:det_opt} is a standard AE reconstruction term computed only for samples with $b[i]=1$. The $\ell_0$ norm in Eq. \eqref{eq:det_opt} counts the number of samples that are included in the reconstruction error; these samples are tagged as ``inliers''. By balancing the reconstruction error and the $\ell_0$ penalty, the hyper-parameter $\lambda$ controls the cost associated with the number of samples used by the AE. 
A large value of $\lambda$ will force the model to include more samples. On the other hand, a small $\lambda$ would lead to a sparser solution with fewer samples included by the model. 
If $\myvec{X}_{in}$ lie near a low dimensional manifold, we assume that the encoder-decoder pair can lead to a good approximation of the inliers, that is $\widehat{\myvec{x}}_i \approx \myvec{x}_i$ for $\myvec{x}_i\in \myvec{X}_{in}$. 
On the other hand, if outliers do not lie near the low dimensional manifold, we expect $\| {\myvec{x}}_i-\widehat{\myvec{x}}_i \|^2_2$ to be large. Unfortunately, due to the $\ell_0$ norm in Eq. \eqref{eq:det_opt} the problem becomes intractable even for a small number of samples. 
To overcome this limitation, following \cite{yamada2018feature,lindenbaum2020deep,lindenbaum2020differentiable}, we propose to replace the deterministic search over the values of the indicator vector $\myvec{b}$ with a probabilistic counterpart.
\subsection{Probabilistic Autoencoder}\label{sec:method}

\begin{figure*}
    \centering
    \includegraphics[width=0.9\textwidth,height=0.27\textheight]{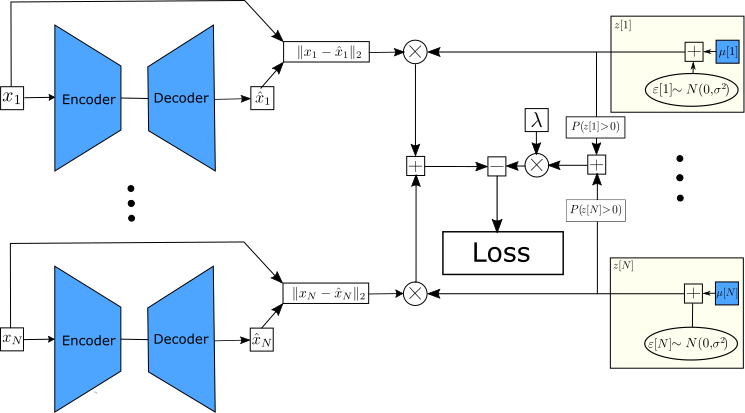}
    \caption{A schematic of the PRAE-$\ell_0$ algorithm (see Eq. \eqref{eq:prob_opt_P}). The $\myvec{x}_i$s (on the left) are the input samples. For any choice of encoder, decoder, and $\myvec{\mu}$ (all in blue), a reconstruction loss is computed (middle) with a subtracted term designed to encourage the model to select only the inlier samples (see Eq. \ref{eq:prob_opt_P}). We optimize over the ``blue" variables (one encoder, one decoder, and a vector $\myvec{\mu}$ that result in the lowest loss). The resulting value ${\mu}[i]$ serves as the outlier score for data point $\myvec{x}_i$.}
    \label{fig:alg_description}
\end{figure*}

We are now ready to present our probabilistic formulation for a sparse AE. 
Towards this goal, we multiply the samples by stochastic gates that relax the binary nature of the indicator vector $\myvec{b}$. The gates are differentiable and are purposed to select a subset of samples on which the AE reconstruction error is minimized. We parameterize a stochastic gate (STG) using mean shifted truncated Gaussian distribution. Specifically, we denote the STG random vector as $\myvec{z} \in [0,1]^N$, parametrized by $\myvec{\mu}\in\mathbb{R}^N$. Each vector entry is defined as
\begin{equation}\label{eq:stg}
  {z}[i] = \max(0,
 \min(1, \mu[i] + \epsilon[i] )) , 
\end{equation} where $\epsilon[i]$ is drawn from  $\mathcal{N}(0 ,\sigma^2)$, $\sigma$ is fixed throughout training, and $\mu[i]$ is a trainable parameter which controls the distribution of the random variable ${z}[i]$. 

We can now incorporate the STGs into our proposed probabilistic AE loss. 
Formally, using the reconstruction loss of ~\eqref{eq:recon_loss}, a probabilistic AE loss can be described using one of the following terms

\begin{equation}\label{eq:prob_opt_P}
L_{p_0}(\myvec{\psi},\myvec{\rho},\myvec{\mu}) = \mathbb{E} \Big(  \sum_{i}  z[i] \Big\| \myvec{x}_i - \widehat{\myvec{x}}_i\Big\|_2^2 -\lambda \| \myvec{z}\|_0 \Big), 
\end{equation}

\begin{equation}\label{eq:prob_opt_E}
    L_{p_1}(\myvec{\psi},\myvec{\rho},\myvec{\mu})= \mathbb{E} \Big(  \sum_{i}  z[i] \Big\| \myvec{x}_i - \widehat{\myvec{x}}_i\Big\|_2^2 -\lambda \| \myvec{z}\|_1 \Big), 
\end{equation}

where, $\lambda$ is a regularization parameter that controls the cost associated with the number of samples included by the AE. To minimize the new loss functions \eqref{eq:prob_opt_P} or \eqref{eq:prob_opt_E}, we propose the following strategy. 
Given some initial guess for the encoder, decoder, parameterized via the weights of $\myvec{\psi} \text{ and } \myvec{\rho}$, we draw realizations for the random vector $\myvec{z}$ and compute the loss value. We note that the regularization terms $\mathbb{E}(\|\myvec{z}\|_0) = \sum P(z[i]>0)$ and  $\mathbb{E}(\|\myvec{z}\|_1) = \sum \mathbb{E}(z[i])$ are parametric, and the expected value of the left term of Eqs. \eqref{eq:prob_opt_P} and \eqref{eq:prob_opt_E} is approximated using Monte Carlo sampling. Then, we differentiate the loss using Stochastic Gradient Descent to update the weights in $\myvec{\psi} \text{ and } \myvec{\rho}$, and the vector $\myvec{\mu}$. A schematic illustration of this procedure is presented in Figure \ref{fig:alg_description}. After convergence, we use the values of the trainable vector $\myvec{\mu}$ as anomaly scores for each data point. A smaller value indicates that the sample should be excluded from the reconstruction loss and therefore is more anomalous. In Section \ref{sec:hyper} we propose two unsupervised schemes for tuning the regularization parameter $\lambda$.


\section{Related Work}

The problem of anomaly detection has been previously addressed using AEs. 
A straightforward solution is to train a standard AE and use the reconstruction error of each sample to quantify if it is normal or anomalous \cite{sakurada2014anomaly}. 
Since this approach does not include regularization, the AE may overfit the outliers and learn a mapping that does not correctly characterize the normal samples. 
To solve this limitation, in \cite{chen2017outlier} the authors propose an ensemble of AEs for anomaly detection. The idea is to train many AEs, each pruned by randomly subsampling the learned connectivities. Then, an aggregated prediction of the ensemble is used to identify the anomalies. One disadvantage of this approach is that it requires extensive computational and memory costs since it involves training hundreds of AEs. Furthermore, the proposed scheme outperforms the ensemble of AEs on several benchmark datasets (see results in supplemental Section S3).

Perhaps the most related method to our work is \cite{zhou2017anomaly}. In \cite{zhou2017anomaly}, the authors proposed the following regularized AE objective
\begin{eqnarray*}
\|\myvec{L}_D -\myvec{\psi}(\myvec{\rho}(\myvec{L}_D))\|_2+\lambda \|\myvec{S}\|_{2,1}\quad
s.t. \quad \myvec{X}-\myvec{L}_D-\myvec{S}=0.
\end{eqnarray*}
This model aims to split the data $\myvec{X}$ into two parts, $\myvec{L}_D$ and $\myvec{S}$ while minimizing the reconstruction error on $\myvec{L}_D$. The regularization in the form of an $\ell_{2,1}$ norm is designed to sparsify the rows (samples) or columns (features) of $\myvec{S}$. This way, the data is split into inliers $\myvec{L}_D$ and a sparse set of outliers $\myvec{S}$. 
To minimize the objective with the additional constraint, they use the Alternating Direction Method of Multiplies (ADMM) \cite{boyd2004convex} with an element-wise projection approach to enforce the constrain.
This method differs from our approach significantly since the regularization relies on the $\ell_{2,1}$ norm applied to $\myvec{S}$. The $\ell_{2,1}$ norm leads to shrinkage of values, and therefore $\myvec{S}$ is not guaranteed to reflect actual samples from $\myvec{X}$. Furthermore, it relies on a different optimization scheme and is not amenable to parallelization through small batch training (due to the additional constraint). Our model overcomes these limitations and, as demonstrated in section \ref{sec:results} leads to more accurate identification of outliers when applied to real and synthetic data. 

\section{Analysis}\label{sec:analysis}
In this section, we justify the use of our proposed probabilistic AE (see Section \ref{sec:method}) to solve the robust auto encoding problem (see Section \ref{sec:sparse}). Since the latter is not differentiable while the first is, our goal is to show that both minimization problems lead to the same solution. 

First, to avoid divergence of the values of $\myvec{\mu}$ in the theoretical analysis, we bound the values of $\myvec{\mu}$ by 
\begin{equation}\label{eq:defM}
    -M\leq \myvec{\mu}[i]\leq M,
\end{equation}
for some large number $M$. Note that the number $M$ is used only for the theoretical analysis and has no practical use when running the algorithm. 

For any vector $\myvec{b} \in \{0,1\}^N$, we define $\myvec{\mu_b}$, such that $\myvec{\mu_b}[i] = -M$ if $\myvec{b}[i] = 0$, and $\myvec{\mu_b}[i] = M$ if $\myvec{b}[i] = 1$ for $i=1\ldots N$. For any $\myvec{\mu}$ we define $\myvec{b}_{\mu}$ such that $\myvec{b}_\mu[i] = sign(\myvec{\mu}[i])$.

We now turn our attention to show that the deterministic optimization problem \eqref{eq:det_opt} (which is not differentiable) is equivalent to our probabilistic optimization \eqref{eq:prob_opt_E} in the following sense
\begin{Theorem}
For any dataset $\myvec{X}$, denote by $(\myvec{\psi}_d, \myvec{\rho}_d ,\myvec{b}_d)$ the minimizer of~\eqref{eq:det_opt} and by $(\myvec{\psi}_p, \myvec{\rho}_p, \myvec{\mu}_p)$ the minimizer of \eqref{eq:prob_opt_E}. Assume that the minimizer of~\eqref{eq:det_opt} is unique and that 
\begin{equation}\label{eq:S_d_gap}
\min\limits_{(\myvec{\psi}, \myvec{\rho} ,\myvec{b}) \neq (\myvec{\psi}_d, \myvec{\rho}_d ,\myvec{b}_d)} L_d(\myvec{\psi}, \myvec{\rho} ,\myvec{b}) \geq L_d(\myvec{\psi}_d, \myvec{\rho}_d ,\myvec{b}_d) + \eps_0
\end{equation}
for some $\eps_0 >0$.
Then for a sufficiently large $M>0$ (see \eqref{eq:defM}), $(\myvec{\psi}_d,\myvec{\rho}_d) = (\myvec{\psi}_p,\myvec{\rho}_p)$, and for any $i = 1,\ldots, L$, $b[i] = 1$ if $\mu[i]>0$ and $b[i] = 0$ otherwise.
\end{Theorem}
In other words if the minimizer of~\eqref{eq:det_opt} is unique then, the encoder, decoder that minimize~\eqref{eq:det_opt} and \eqref{eq:prob_opt_E} are equivalent. Moreover, the samples included by both models (indicated by $\myvec{b}$ and $\myvec{z}$) are the same.
\begin{proof}

The proof construction is comprised of three arguments. The final argument relies on the first two and concludes the proof.

\textbf{Argument 1:} 
For any triplet $(\myvec{\psi}_d,\myvec{\rho}_d,\myvec{b})$ the deterministic loss $L_d$ can be approximated by the probabilistic loss $L_{p_1}$. Namely, for any $\eps,\delta>0$ there is a value of $M>0$ such that 
\[
|L_d(\myvec{\psi}_d, \myvec{\rho}_d ,\myvec{b}_d) - L_{p_1}(\myvec{\psi}_d,\myvec{\rho}_d,\myvec{\mu_b})| \leq \eps,
\]
with probability $1-\delta$.

To prove this argument, we first compute the expectation $E(z)$ using definition \eqref{eq:stg}, we get: 
\begin{align*}
E(z) = &\mu 
- \frac{1}{\sqrt{2\pi}} \int_{-\infty}^{0} t e^{-\frac{(t-\mu)^2}{2\sigma^2}}dt 
- \frac{1}{\sqrt{2\pi}} \int_{1}^{\infty} t e^{-\frac{(t-\mu)^2}{2\sigma^2}}dt \\
&+ \frac{1}{\sqrt{2\pi}} \int_{1}^{\infty}  e^{-\frac{(t-\mu)^2}{2\sigma^2}}dt,     
\end{align*}

computing the integrals leads to:
\begin{align*}
E(z) &= \frac{\sigma}{\sqrt{2\pi}} 
       ( e^{-\frac{\mu^2}{2\sigma^2}} - e^{-\frac{(1-\mu)^2}{2\sigma^2}}) +
    (\mu-1)* \Phi(\frac{1-\mu}{\sigma}) \\&- \mu*\Phi(-\frac{\mu}{\sigma}) + 1,
\end{align*}
where $\Phi$ is the CDF of the standard normal distribution.

Since $\lim_{\mu \to \infty} E(z) = 1$, and $\lim_{\mu \to -\infty} E(z) = 0$, than, for any $\eps>0$, there is a sufficiently large $M$, such that
\begin{equation}\label{eq:lambda_diff}
\left| \lambda \sum_i {\mathbb{E}}(z[i]) -\lambda \|\myvec{b}\|_0  \right| <\eps/2
.    
\end{equation}

From the definition of $z$ we also know that for $\mu>1$, $P(z\neq 1) = \Phi(\frac{1-\mu}{\sigma})$, and thus $\lim_{\mu \to \infty} P(z = 1) = 1$. Similarly $\lim_{\mu \to -\infty} P(\textnormal{z} = 0) = 1$. Thus, for any $\delta$ there is $M$ large enough, such that

\begin{equation}\label{eq:approx_diff}
\left| \sum_i z[i]\|\myvec{x}_i - \hat{\myvec{x}}_i \|^2_2 -  \sum_i {b}[i]\|\myvec{x}_i - \hat{\myvec{x}}_i\|^2_2   \right| <\eps/2,
\end{equation}
with probability $1-\delta$.

Combining \eqref{eq:lambda_diff} and \eqref{eq:approx_diff}, we have that for any $\delta > 0$, there is a value of $M$ such that 
\[
|L_d(\myvec{\psi}_d, \myvec{\rho}_d ,b_d) - L_{p_1}(\myvec{\psi}_d,\myvec{\rho}_d,\myvec{\mu_b})| \leq \eps,
\]
with probability $1-\delta$. This concludes the proof of Argument 1.

\textbf{Argument 2:} For any AE $(\myvec{\psi},\myvec{\rho})$, the minimum $\min_{\myvec{\mu}} L_{p_1}(\myvec{\psi},\myvec{\rho},\myvec{\mu})$ is achieved when $\myvec{\mu}[i]$ equals to either $M$ or $-M$ for all $i$. 

Assume by contradiction that the minimum of $L_{p_1}$ is achieved at a point where for some $k$, $\myvec{\mu}[k]$ is not either $M$ or $-M$. If 
 $
 \| \myvec{x}_i - \widehat{\myvec{x}}_i\|_2^2 \geq \lambda,
$
then for $\hat{\myvec{\mu}}$ such that  $\hat{\myvec{\mu}}[i] = \myvec{\mu}[i]$ for all $i \neq k$ and $\hat{\myvec{\mu}}[k] = -M$ we have that  
$ 
L_{p_1}(\myvec{\psi},\myvec{\rho},\hat{\myvec{\mu}}) \leq L_{p_1}(\myvec{\psi},\myvec{\rho},\myvec{\mu}),
$ 
which contradicts the minimality of  $L_{p_1}(\myvec{\psi},\myvec{\rho},\myvec{\mu}) $.
In case 
$
 \| \myvec{x}_i - \widehat{\myvec{x}}_i\|_2^2 \leq \lambda
 $
 a similar argument will lead to a contradiction as well.
 
 \textbf{Argument 3:} Assume by contradiction that the minimizers of \eqref{eq:det_opt} and \eqref{eq:prob_opt_E} are not equivalent, i.e.
 \begin{equation}\label{eq:assume_contredict}
     (\myvec{\psi}_d, \myvec{\rho}_d ,\myvec{\mu}_{b_d}) \neq (\myvec{\psi}_p, \myvec{\rho}_p ,\myvec{\mu}_p).
 \end{equation}
From Argument 2 we have that $\mu_p[i] = M \mbox{ or } -M$ for all $i$. 
From Argument 1 we have that 
\begin{align}  
\label{eq:dist_Sp_Sd_minimal}
\|L_d(\myvec{\psi}_p,\myvec{\rho}_p,\myvec{b}_{\mu_p}) - L_{p_1}(\myvec{\psi}_p,\myvec{\rho}_p,\mu_p)\| &\leq \eps ~\mbox{and}\nonumber
\\
\|L_d(\myvec{\psi}_d,\myvec{\rho}_d,\myvec{b}_{d}) - L_{p_1}(\myvec{\psi}_d,\myvec{\rho}_d,\mu_{b_d})\| &\leq \eps.
\end{align}
Since $ \myvec{\psi}_p,\myvec{\rho}_p, \myvec{\mu}_{p}$ is the minimizer of $L_{p_1}$, we have from \eqref{eq:dist_Sp_Sd_minimal} that  
\begin{equation}\label{eq:dist_b_d_b_mu}
    L_{d}(\myvec{\psi}_d,\myvec{\rho}_d,\myvec{b}_{d}) \geq  L_d(\myvec{\psi}_p,\myvec{\rho}_p,b_{\mu_{p}}) - 2\eps.
\end{equation}
From Eq. \eqref{eq:assume_contredict} and the assumption of the theorem in Eq. \eqref{eq:S_d_gap}, we have that 
\begin{equation}\label{eq:dist_Sb_Sbp_minimal}
\|L_d(\myvec{\psi}_p,\myvec{\rho}_p,\myvec{b}_{\mu_p}) - L_d(\myvec{\psi}_d,\myvec{\rho}_d,\myvec{b}_d)\| \geq \eps_0.
\end{equation}
For $2\eps<\eps_0$, Eq. \eqref{eq:dist_Sb_Sbp_minimal} contradicts Eq. \eqref{eq:dist_b_d_b_mu}. 
\end{proof}




\begin{figure}[tb!]
    \centering  
    \includegraphics[width=0.5\textwidth]{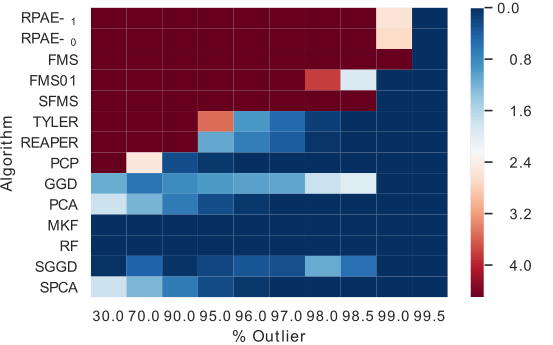}
    \caption{Comparing PRAE to several Robust Subspace Recovery (RSR) algorithms. The y-axis represents the different algorithms, and the x-axis represents different percentiles of outliers. Each box is colored according to the mean over $10$ runs of the log of the angle between the recovered subspace and the ground truth.}
    \label{fig:RSR_comparison}
\end{figure}

\section{Experiments} \label{sec:results}
\subsection{Robust Subspace Recovery Problem}\label{sec:lin}
First, we test the performance of the proposed algorithms in the linear setting. While this regime has fewer applications, it is well studied, and it is easier to analyze and compare different methods.

We note that in the linear regime, the outlier detection problem is strongly related to the Robust Subspace Recovery problem (RSR). Thus, this section focuses on comparing our proposed scheme to baseline methods designed to solve the RSR problem. The RSR problem involves finding a low-dimensional (linear) subspace in a corrupted, potentially high-dimensional dataset. For a complete overview of RSR, we refer the reader to \cite{lerman2018overview}.

Following \cite{lerman2018overview}, for any chosen persentile of outliers $r = N_{out}/N$, we generate $N = 10000$ points in $\RR^{200}$ in the following way:
first we randomly generate $\myvec{X}_{in}^{low}$, a set of $N_{in} = (1-r)N$ random points in $\RR^{10}$. Next we generate a random linear transformation $\myvec{T}\in \RR^{200\times 10}$, and set $\myvec{X}_{in}^{high}= \myvec{T} \myvec{X}_{in}^{low}$. Finally, we generate $\myvec{X}_{out}$ as $N_{out}=rN$ random points in $\RR^{200}$, and define the dataset $\myvec{X} = \myvec{X}_{in}^{high} \cup \myvec{X}_{out}$. The task is to recover $\myvec{T}$ and $\myvec{X}_{in}$ given the data $\myvec{X}$. The accuracy is measured by the $\log$ of the angle between the recovered $\myvec{T}$ and the correct $\myvec{T}$. Each experiment was performed $10$ times, and the final outcome is the average of the $10$ runs. For complete implementation details we refer the reader to supplemental Section S1.

The result of the comparison to other algorithms under different percentile of outliers ($r = N_{out}/N$) appears in Figure~\ref{fig:RSR_comparison}. We compared to leading schemes form the evaluation in \cite{lerman2018overview}, namely:
fast median subspace (FMS) \cite{lerman2018fast}, 
Tyler’s M-estimator (TYLER) \cite{zhang2016robust},
REAPER \cite{lerman2015robust},
the augmented Lagrange multiplier method (PCP)\cite{lin2010augmented},
geodesic gradient descent (GGD) \cite{maunu2019well},
and principal component analysis (PCA).


While our approach is not explicitly designed for the RSR problem, it is easy to see that our algorithms perform on par with state-of-the-art methods for RSR. Even for 99\% outliers, in 7 out of 10 runs, our algorithms found exactly all the inliers. 
We note that since our approach is not designed for RSR and focused on the more general non-linear setting, FMS recovers a more accurate subspace and requires a shorter training time. 
Nonetheless, we argue that this experiment highlights that our model is relatively robust to the number of outliers. We observe that PRAE can correctly recover inliers in a noisy setting. Precisely, when we use $\myvec{X}_n = \myvec{X} + \eta$, where the noise $\eta \sim N(0,10^{-2}I)$ and 99\% outliers, our model is still able to correctly identify a subset of inliers that are sufficient for subspace recovery in 7 out of 10 cases. However, we omit this noisy RSR experiment since it requires a more involved method for reconstructing the subspace-based on the (correctly) retrieved noisy inliers.



\begin{figure*}[htb!]
    \centering  
    \includegraphics[width=0.40\textwidth]{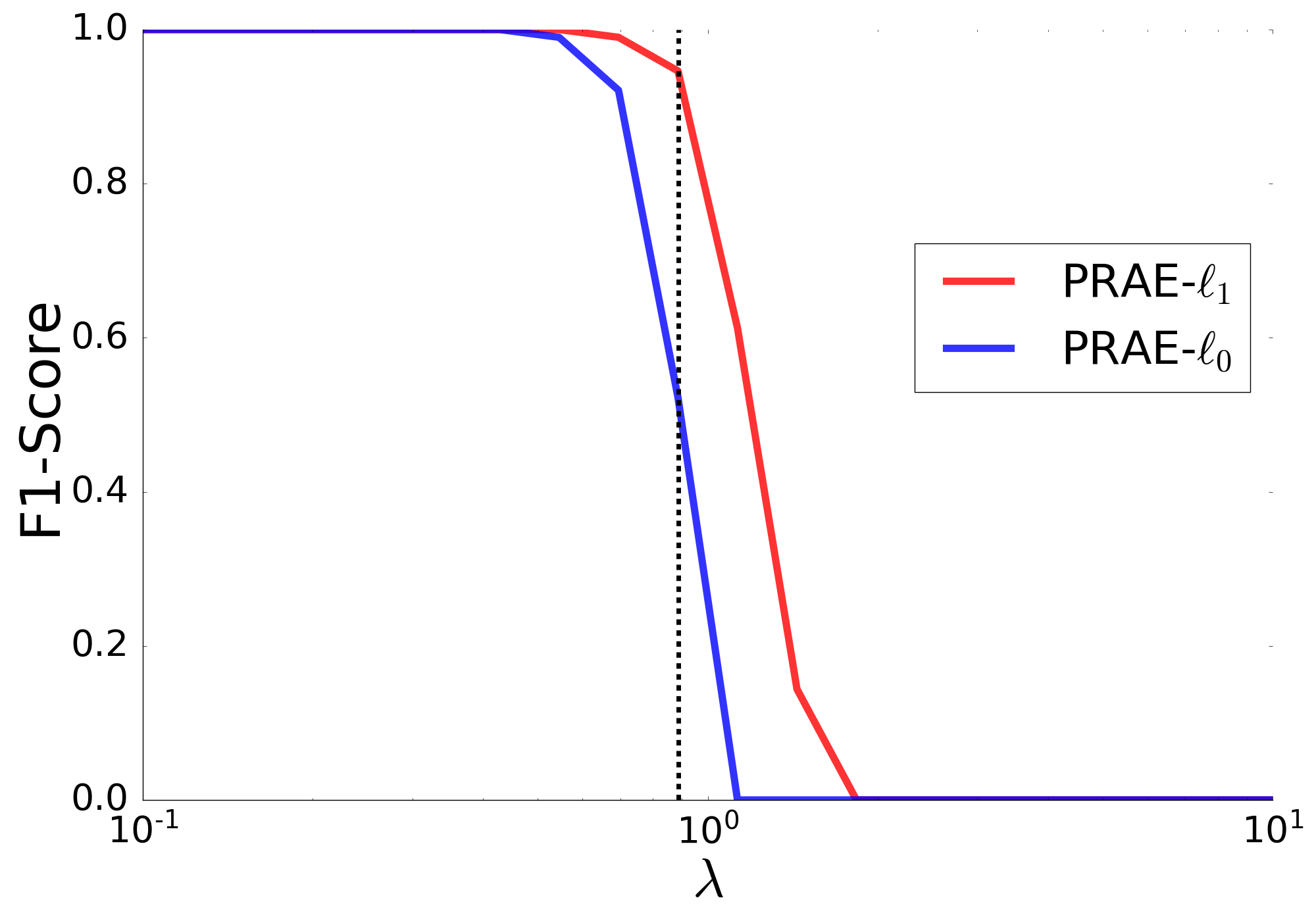}~
    \includegraphics[width=0.40\textwidth]{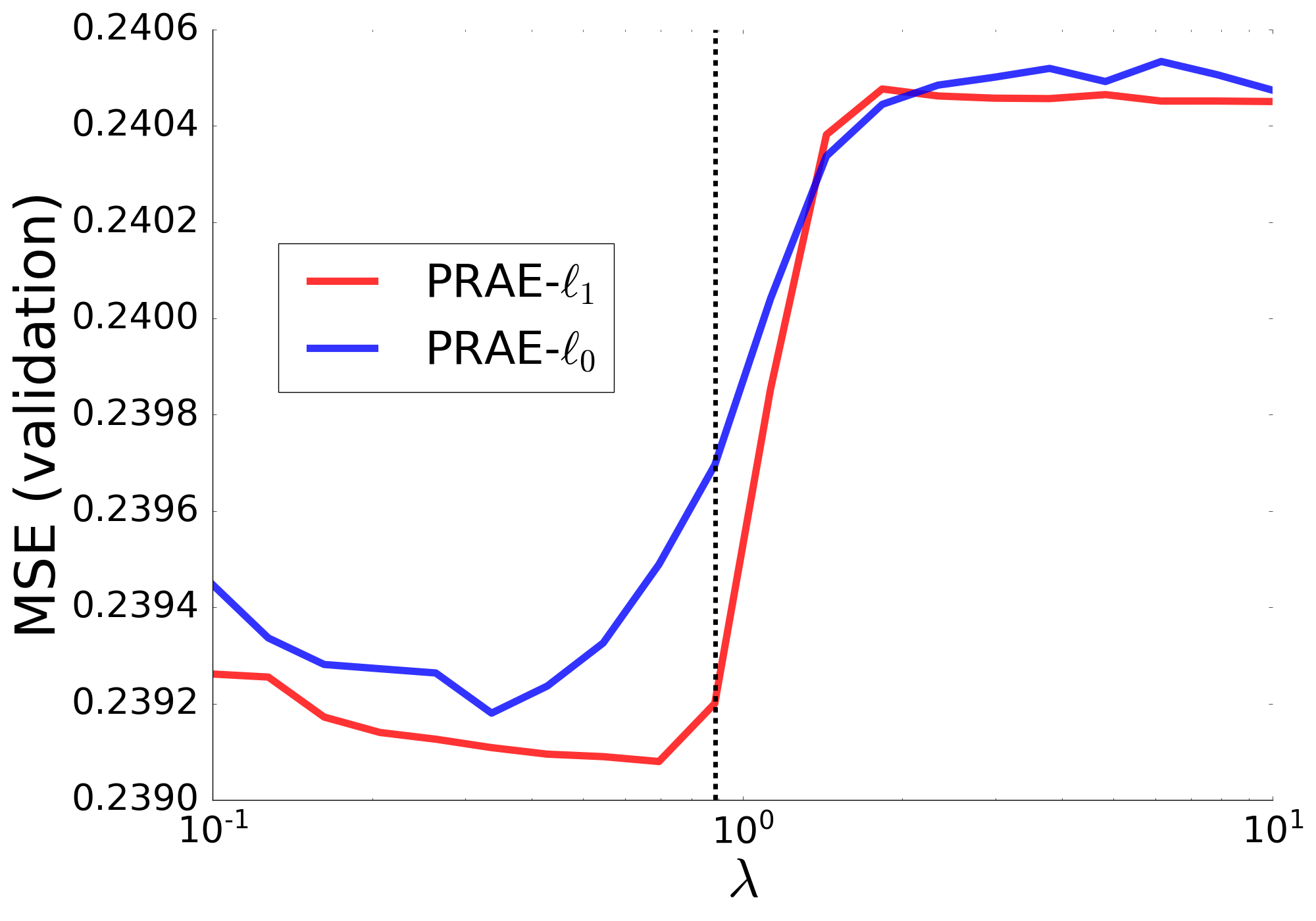}
    \caption{Phase transition of PRAE. As we increase $\lambda$ above a certain threshold, PRAE starts to include outliers resulting in a lower F1-score (left panel) and larger reconstruction error (MSE) on unseen samples (right panel). Dashed line indicates our (unsupervised) estimation of the value of $\lambda$ in which the proposed scheme transitions from removal to inclusion of outliers.  }
    \label{fig:phase}
\end{figure*}

\subsection{Unsupervised Schemes for Tuning the Regularization Parameter}\label{sec:hyper}
One practical consideration in PRAE is the choice of regularization parameter $\lambda$. In this section, we empirically study the effect of this parameter and propose two unsupervised schemes for tuning $\lambda$. Then, we use synthetic data to demonstrate that our estimated value of $\lambda$ leads to accurate identification of all inliers and removal of all outliers.  

We focus on the linear data model described in section \ref{sec:lin}, but with $N=200$, $\myvec{X}^{low}_{in}\in \RR^{150 \times 2}$, and $\myvec{T} \in \RR^{100\times 2}$. We generate data from this model and run PRAE-$\ell_0$, and PRAE-$\ell_1$ for various values of $\lambda$ in the range $[0.1,10]$.  We run each model $20$ time and record the average F1-score which is computed based on precision and recall of outlier identification. Specifically, we define an outlier $\myvec{x}_i$ as a sample such that after training $\mu[i]<thresh$, and an inlier otherwise. Here, $\mu[i]$ is computed based on Eq. \ref{eq:stg} but without the injected Gaussian noise. We set $thresh$ to $0.1$ although other values within $(0,1)$ lead to similar results.

 In both proposed loss functions (see \eqref{eq:prob_opt_P} and \eqref{eq:prob_opt_E}) $\lambda$ balances between the number of sample included by the model and the reconstruction loss. For a very large $\lambda$, we expect the model to include all samples since the regularization term would be larger than the reconstruction of $\myvec{x}_i$ (for inliers and outliers). On the other hand, if $\lambda=0$ all samples should be excluded by the model. For small values of $\lambda>0$, we expect the model to include the inliers (since we can obtain zero reconstruction loss) and exclude the outliers. Based on the linear model experiment (described above), we observe a "phase transition" in the behavior of PRAE as a function of $\lambda$. Namely, as evident in the left panel of Figure \ref{fig:phase} for small values of $\lambda$, PRAE accurately removes all outliers and includes all inliers. 
 
 In this example, since all samples have roughly the same energy ($\ell_2$ norm), we can propose a simple scheme for estimating the $\lambda$ value in which the phase transition occurs. Specifically, we can compute the mean energy of all samples, namely $ME=\frac{1}{N}\sum^N_i \|\myvec{x}_i\|^2_2$. Since we can not reliably reconstruct the outliers (based on our data model), we expect the error for reconstructing outliers to be $\sim ME$. Therefore, for any $\lambda>ME$, PRAE-$\ell_1$ should include outliers since $\| \myvec{x}_i-\hat{\myvec{x}}_i\|^2_2$ is compared to $\lambda$ in loss (see \eqref{eq:prob_opt_E}). On the other hand, if  $\lambda<ME$, PRAE-$\ell_1$ should exclude outliers (based on the same argument). For PRAE-$\ell_0$, this argument is not precise; nonetheless, we observe that ME lines well with the phase transition of both models. This is presented as a dashed black line in Figure \ref{fig:phase}.

Another scheme to tune the regularization parameter is based on the value of $\lambda$ that minimizes the reconstruction loss of unseen samples (a validation set). Here, we assume that inliers can be represented by a low dimensional space while outliers can not. By evaluating the model's reconstruction error on unseen samples, we can check if the model suffered from overfitting on anomalies or has used only inliers. We repeat the experiment above but evaluate the average reconstruction loss on $200$ unseen samples generated from the same model. As evident in the right panel of Figure \ref{fig:phase} both models lead to the smallest reconstruction loss for $\lambda$ values that coincide with the perfect F1-score (supporting the validity of the proposed tuning scheme). We observe that PRAE-$\ell_0$ leads to a higher reconstruction error for large values of $\lambda$. This might indicate that the inclusion of all samples occurs earlier in training leading to stronger overfitting. 

\subsection{Evaluation Based on Synthetic Data }\label{sec:syth}
To illustrate the qualities of our scheme in a nonlinear setting, we use a simple artificial example suggested by \cite{lai2019robust}. For the normal samples, we consider a ``narrow swiss-roll'', with $1000$ points uniformly sampled from $ [ 3\pi/2, 9\pi/2] \times [0,0.1]$, and embedded into $\mathbb{R}^3$ using $(t,h)\rightarrow (t\cos(t),h,t\sin(t))$. 
Then, we generate additional $200$ ``outliers'' sampled from $N(0,\sigma_N^2 I_3)$, where $I_3\in \RR^{3\times 3}$ is the identity matrix. 
We generate such data with several values of $\sigma_N^2$, and following \cite{chen2017outlier,ishii2019low} we evaluate the quality of multiple anomaly detection methods using Receiver Operating Characteristic (ROC)
curves. The ROC measures the trade-off between true positive and false positive rates. The true positive rate is defined as the ratio between identified anomalies and true anomalies, while the false positive rate is the portion of normal samples identified as anomalies. The ROC curves are summarized by measuring the AUC (area under the curve). 
We compare our method to several strong baselines with code available at~\cite{zhao2019pyod}. 
Specifically, we use CBLOF \cite{cblof} a local clustering-based approach, COF~\cite{COF} which uses the density of the data, SOD \cite{SOD} and HBOS \cite{HBOS} which are based on proximity, IForest \cite{Iforest} and LSCP \cite{lscp} which are ensemble methods. We also compare to the probabilistic ABOD \cite{abod} and to OC-SVM \cite{scholkopf2001estimating}. 
Finally, we compare to several NN based methods, such as: $\ell_{2,1}-AE$ \cite{zhou2017anomaly}, Deep-SVDD \cite{ruff2018deep}, and RSR-AE \cite{lai2019robust}. 
Table \ref{tab:non_lin_synth} shows the median AUC of all baselines over ten runs. For all NN models (including PRAE), we use five hidden layers and a latent dimension of $2$. 
Since the inliers lie near a manifold (``swiss-roll'') with an intrinsic dimension of $2$, the AE can correctly capture the swiss-roll's mapping. 
As indicated by table \ref{tab:non_lin_synth}, PRAE accurately distinguishes inliers from outliers for several scales of $\sigma_N^2$.

\begin{table*}[htb!]
  \centering
  \begin{adjustbox}{width=1\textwidth,center}
    \begin{tabular}{lccccccccccccc}
     \toprule
    $\sigma_N^2$ & \multicolumn{1}{l}{CBLOF} &  \multicolumn{1}{l}{ABOD} & \multicolumn{1}{l}{COF } & \multicolumn{1}{l}{IForest} & \multicolumn{1}{l}{SOD} & \multicolumn{1}{l}{LSCP} & \multicolumn{1}{l}{HBOS}& \multicolumn{1}{l}{OC-SVM} & \multicolumn{1}{l}{DSVDD} &\multicolumn{1}{l}{RSR-AE}  &  $\ell_{2,1}$-AE&  \multicolumn{1}{l}{PRAE-$\ell_0$} & \multicolumn{1}{l}{PRAE-$\ell_1$}\\
         \bottomrule

    0.1  & 13.81 &48.77 & 63.39   &  26.08 & 31.54 & 67.72& 60.65 & 68.54 &69.19  & 93.12& 68.47  & \bf{96.23}& \color{blue}{\bf{97.48}}\\
    1 & 28.88 &  54.67 &  64.64 & 55.20 & 76.35 & 67.99& 81.34 & 50.58 & 71.97& 82.25 & 46.23 &  \color{blue}{\bf{99.97}} & \bf{99.94} \\
    10  & 93.77 & 53.01 &  98.55  & 97.87 & 93.28 & 99.65 &94.34 & 52.81& 40.75&  75.13& 80.59 &   \color{blue}{\bf{99.55}} &  {\bf{99.42}}  \\

     \bottomrule
    \end{tabular}%
      \end{adjustbox}
        \caption{Performance comparison on a synthetic dataset. Each value corresponds to the median AUC of the ROC curve of a different algorithm (column) under different anomaly variance (row). Blue and bold indicate first and second ranked methods respectively.}
  \label{tab:non_lin_synth}%
\end{table*}%

\subsection{Anomaly Detection on Real Data} 
Next, we evaluate the proposed approach on a diverse set of real-world datasets whose properties appear in the right column of table \ref{tab:realresults}. Since we focus on unsupervised anomaly detection, we wish to identify anomalies in two settings: 
(1) ``in-sample'', in which we aim to curate a given dataset from outliers, and 
(2) ``out-of-sample'', in which we want to identify outliers from newly arrived samples without additional training. 
To evaluate performance on (1), we apply our algorithm to the entire dataset and use the value of the gate parameter $\mu[i]$ as an anomaly score for $x_i$. 
For (2), we split each dataset and use one half for training the PRAE while the hold-out-set is used to evaluate the performance without additional training.
Here, since we don't have gates from unseen samples, the anomaly score for $\myvec{x}_i$ is based on the reconstruction error $\|\myvec{x}_i-\widehat{\myvec{x}}_i\|^2_2$.

\begin{table*}[htb!]
  \centering
  \begin{adjustbox}{width=1\columnwidth,center}
    \begin{tabular}{lcccccccccccccc}
     \toprule
    dataset & \multicolumn{1}{l}{CBLOF} &  \multicolumn{1}{l}{ABOD} & \multicolumn{1}{l}{COF } & \multicolumn{1}{l}{IForest} & \multicolumn{1}{l}{SOD} & \multicolumn{1}{l}{LSCP} & \multicolumn{1}{l}{HBOS}& \multicolumn{1}{l}{OC-SVM} & \multicolumn{1}{l}{DSVDD} &\multicolumn{1}{l}{RSR-AE}  &  $\ell_{2,1}$-AE&  \multicolumn{1}{l}{PRAE-$\ell_0$} & \multicolumn{1}{l}{PRAE-$\ell_1$} &\multicolumn{1}{l}{$N/D/$ Out \% }\\
         \bottomrule
    Lympho &   95.12  &  72.53 &  88.08 & \color{blue}{\bf{99.98}} & 92.08 & 97.38 & \bf{99.01} &97.88 & 81.13 & 63.90& 98.72 & 96.09 & 95.98 &148/ 18/ 4.1\\
        Ecoli  & 87.90 & 75.72 &  87.77  & 86.41 & 82.67 & 86.71 & 81.44 & 86.71& 92.07&  76.80& 44.21 &   \bf{88.94} &  \color{blue}{\bf{89.09}} &336/ 7/ 2.7 \\
            Cardio  & 85.31 &  49.87 &  51.79 & 92.15 & 69.19 & 59.42& 88.14 & 58.39 & 61.89& 69.44 & 88.10 &  \color{blue}{\bf{94.28}} & \bf{93.87} & 1831/ 21/ 9.6 \\
           Yeast & 68.63 &  56.81 & 53.18 & 79.30 & 61.79 & 61.62 & 78.98 & 62.16& 66.62& 79.54& 70.58 & {\bf{83.37}} & \color{blue}{\bf 83.95} &1364/ 8/ 4.8 \\
              Musk & 95.17 & 68.04 & 52.31 & 97.46 & 89.93& 60.45&98.91  & 90.28& 80.17&82.59 & \bf{98.96} &\color{blue}{\bf{99.17}} & 98.61 & 3062/ 166/ 3.1\\
    Thyroid  & 91.68 &48.73 & 58.34   &  \color{blue}{\bf{97.64}} & 87.57 & 81.91& \bf{95.06} & 84.38 &90.09  & 90.12 & 81.39  & 94.78 & 94.69 &3772/ 6/ 2.1\\
    MNIST-S & 52.97 & 67.67 & 69.93 & 83.37 & 70.64 & 92.18 &77.09 & 90.20& 85.01 & 91.29 & 90.39 &\color{blue}{\bf{93.54}} & {\bf 93.31} &5127/ 784/ 5.2\\
   Fashion-MNIST & 89.04 & 50.65 & 69.25 & 91.79 & 64.74 & 82.96 & 74.75 & 88.87 & 78.28 & 82.86 & 90.69 &{\bf{93.89}} & \color{blue}{\bf 93.99} &5300/ 784/ 5.6\\
   PBMC & 76.36 & 51.50 & 90.71 & 88.71 & 79.48 & NA &70.36 & 88.69 & 55.67 & 51.10 & 64.26 &{\bf{90.91}} & \color{blue}{\bf 91.30} &6300/ 32738/ 4.7\\

    Pendigits & 91.73 & 63.13   & 52.33 & 95.18 & 64.91 & 47.39 & 92.69 & 88.14& 78.82& 89.54& 93.73 &  \color{blue}{\bf{97.56}} & \bf{97.47} & 6870/ 16/ 2.2 \\
 Mammog. & 81.42 & 83.00 & 71.18 & 86.62 &  77.51 & 50.53 & 83.00  & 81.29& 83.76& 82.40& 81.31 &   \bf{88.31} & \color{blue}\bf{88.33} &11183/ 6/ 2.3  \\
    Shuttle & 66.50 & 58.22 & 53.59 & \color{blue}{\bf 99.67 }& 50.74 & 53.29 & 98.42 &63.66 & 97.89&82.31 & 97.78  &{\bf99.13} &  98.95 &49097/ 9/ 7.1\\
 Credit  & NA & NA & NA &  95.01 & NA & NA & 94.36 & \color{blue}{\bf 95.39} & 58.78 &84.57 & 92.86  &{\bf95.25} &  91.31 & 284807/ 16/ 0.17\\

     \hline
     Average-AUC &83.21	&62.51&	67.71	&91.14&  76.41& 70.05& 86.15& 84.36&
       76.02& 78.67& 82.93&	\color{blue}{\bf{93.01}}&	{\bf 93.46}\\
Median rank &6& 12&  12 &  3 &  9& 10&  2&  7&  9 &  7 &  5&
        2 &  2\\
     \bottomrule
    \end{tabular}%
      \end{adjustbox}
        \caption{Performance comparison with several leading baselines. We present the median AUC over $10$ runs. Blue and bold indicate first and second-ranked methods respectively. NA indicates that the baseline did not converge on the dataset. }
  \label{tab:realresults}%
\end{table*}%

We train our proposed AE with an encoder-decoder pair with five hidden layers each of size $10$; the hidden dimension is $1$ (this might not be optimal but worked well across most datasets). We use the heuristic proposed in section \ref{sec:hyper} to tune the regularization parameter to $\lambda=1<ME$. We run all methods $10$ times and record the ROC for each run. 
In table \ref{tab:realresults} we present the median AUC of the proposed method and all baselines evaluated for the ``in-sample'' setting. These results show that the proposed approach compares favorably to leading methods on a wide range of datasets.
Specifically, PRAE outperforms all baselines in terms of average AUC. 
Moreover, both PRAE-$\ell_0$ and PRAE-$\ell_1$ have a median rank of $2$, and each would have a median rank 1 in the absence of the other approach as a competitor. 

In the supplemental material, we further demonstrate that our method can accurately identify outliers in the ``out-of-sample'' setting (Section S2). 
Box plots indicating the stability of our approach also appear in Section S2. We also provide: reproducibility required technical details (Section S3 and S4), hyperparameters sensitivity evaluation (Section S5), ablation study (Section S6), a deeper analysis of MNIST-S and Fashion MNIST (Section S8), and an analysis of the running time and computational complexity of the method (Section S9).

\section{Strengths and Limitations}\label{sec:strength_limitations}
The proposed method provides several advantages compared to a standard autoencoder: (1)
It removes outliers along the training process, therefore, can more accurately identify a low-dimensional subspace that represents the normal samples. 
(2) It provides a reliable metric for curating training data via an "in-sample" fashion anomaly detection. (3) It learns an encoder-decoder mapping that can be used for filtration of anomalies from new unseen samples ("out-of-sample" regime). The success of our method relies on the assumption that normal samples lie near a latent low dimensional manifold, while the outliers are diverse and do not obey such structure. Indeed, there are cases where this assumption does not hold, and our method will not be optimal for identifying the outliers. While robust PCA assumes inliers samples are low rank, our method identifies those as samples that can be ``easily'' reconstructed by an AE. To prevent the AE from overfitting to outliers, we limit its capacity using a low-dimensional bottleneck. Nonetheless, understanding the properties of an AE from an optimization perspective is an open research direction. A related problem has been studied in supervised learning in recent years \cite{tishby2015deep,ronen2019convergence}. Another limitation of our method is the regularization parameter $\lambda$. In our experiments, we kept $\lambda=1$ following the intuition provided in section \ref{sec:hyper}. Nonetheless, a rigorous scheme for finding the optimal value of $\lambda$ is an interesting question for future work. Another interesting direction involves incorporating the proposed probabilistic robust mechanism into an attention architecture. 

\section{Conclusion}
In this work, we present a novel methodology for anomaly detection. Our method, which we call Probabilistic Robust autoencoder (PRAE), is based on a regularized AE designed to remove samples that do not lie near a low dimensional manifold. Specifically, we multiply each data instance with an approximately binary random variable and add a penalty term to the AE training to encourage sparsity of the number of samples used by the model. We prove an equivalence between the solution of our probabilistic formulation and the solution of an intractable $\ell_0$ regularized AE. Finally, we demonstrate different properties of the proposed method using extensive simulations. Overall, we obtain several state-of-the-art results on synthetic and real datasets with diverse properties (size, dimensions, and type of features).

\bibliography{references}

\bibliographystyle{unsrt}

\clearpage

\beginsupplement
\section*{Supplemental Material}

\label{appendix:lin_experiment}
Here, we describe the technical details for the experiment performed in Section 6.1. 
We generate $N = 10000$ points in $\RR^{200}$ in the following way:
first we generate $N_{in}$ points in $\RR^{200}$ with distribution $N(0,I_{200})$ where $I_{200} \in \RR^{200 \times 200}$ is the identity matrix. This set of points is orthogonaly projected into a random $10$ dimensional space and denoted by $\myvec{X}_{in}^{high}$. 
The orthogonal projection is performed by projecting onto the basis of the column space of a random Gaussian matrix. $\myvec{X}_{in}^{high}$ are normalized so that the expectation of the norm is $1$.
Next, we generate $\myvec{X}_{out}$ as random points in $\RR^{200}$ with distribution $N(0,1/\sqrt{200}I_{200})$. This is done so that the expected value of the norm of the outliers will be equal to the expected value of the norm of the inliers.
Finally, the dataset is constructed by combining $\myvec{X}_{out}, \myvec{X}_{in}^{high}$ and adding noise $\sim N(0,\sigma^2)$. For the results in Table 2 
a negligible noise with $\sigma^2 = 10^{-8}$ was added. In the second experiment a noise of $\sigma^2 = 10^{-2}$ was added.

\section{Additional Experimental Details}
\label{appendix:additional_exper}
First, we present the result for the ``out-of-sample'' setting, in which we split each dataset and use one half for training while the hold-out-set is used to evaluate the performance without additional training. Here, since we don't have gates from unseen samples, the anomaly score for $\myvec{x}_i$ is based on the reconstruction error $\|\myvec{x}_i-\widehat{\myvec{x}}_i\|^2_2$. We compare the performance of each of the algorithms that were compared in the ``in-sample'' setting. The results are summarized in Table \ref{tab:realresults_oos}. As evident from this experiment, the proposed approach works well in the ``out-of-sample'' setting. Specifically, the AUC improves on most datasets, and the proposed schemes outperform all baselines in terms of average AUC.

\begin{table*}[htb!]
  \vskip -0.01 in
  \centering
  \begin{adjustbox}{width=1\columnwidth,center}
    \begin{tabular}{lcccccccccccc}
     \toprule
    Data set & \multicolumn{1}{l}{CBLOF} &  \multicolumn{1}{l}{COF } & \multicolumn{1}{l}{IForest} & \multicolumn{1}{l}{SOD} & \multicolumn{1}{l}{LSCP} & \multicolumn{1}{l}{HBOS}& \multicolumn{1}{l}{OC-SVM} & \multicolumn{1}{l}{DSVDD} &\multicolumn{1}{l}{RSR-AE}  &  $\ell_{2,1}$-AE& \multicolumn{1}{l}{PRAE-$\ell_0$} & \multicolumn{1}{l}{PRAE-$\ell_1$}\\
         \bottomrule

    Thyroid  & 90.92 & 68.44   &  \color{blue}{\bf{97.85}} & 88.41 & 93.28& \bf{95.91} & 96.13 &89.74  & 75.31& 95.50 & 95.81 & \bf{95.91} \\
    Cardio  & 84.54 &   54.52 & 91.95 & 67.28& 70.2& 84.10 & 58.39 & 61.89& 69.44 & \bf{94.08}  & \color{blue}{\bf{94.58}} & 93.03 \\
    Ecoli  & 81.48 &  87.77  & 86.41 & 92.50 & 95.60 & \bf{96.84} & \color{blue}{\bf{99.26}} & 92.07&  76.39& 49.57 &  90.82 &  93.36  \\
    Lympho &   95.30  &  98.59 & \color{blue}{\bf{99.06}} & 84.50 & 98.59 & 99.98 &99.28 & 80.20& 63.90& \bf{99.24} &  93.13 & 93.84 \\
    Pendigits & 97.16 &  52.2 & 93.35 & 69.58 & 50.45 & 92.69 &95.31& 78.57 & 76.02& 93.51 &  \color{blue}{\bf{98.72}} & \bf{98.93} \\
       Yeast & 67.39 & 59.09 & 76.52 & 78.89 & 61.62 & 82.76 & 81.70& 72.24& 49.91 & 74.37 & {\bf{81.86}} & \color{blue}{\bf 83.48} \\
   Musk & 95.29 &  54.77 &97.48 & 93.68& 92.27&99.28  & 70.92& 80.17&77.76 & \bf{99.99}  &\color{blue}{\bf{98.95}} & 99.74 \\
 Mammography & 83.94 &  71.98 & 86.26 &  79.67 &51.90 & 83.39  & 87.48& 83.76& 82.20& 81.59 &   \bf{88.10} & \color{blue}\bf{88.32}  \\
    Shuttle & 68.75 &  51.45 & \color{blue}{\bf 99.97 }& 58.28 & 53.76 & 98.43 &99.14 & 97.89&71.61 & 98.81 &{\bf99.22} &  99.10 \\
    MNIST-S & 52.97 &  52.23 & 83.37 & 51.14 & 51.57 & 51.36 & 90.20& 80.92 & 91.29 & 90.19  &\color{blue}{\bf{93.54}} & {\bf 93.31}\\
 Fashion-MNIST & 70.69 & 78.11 & 92.12 & 88.54 & 88.12 & 75.64 & 88.10 & 86.66 & 68.25 & 89.77  &\color{blue}{\bf{90.32}} & {\bf 89.18}\\
  PBMC & 90.28 &  92.29 & 83.37 & 90.28 & 91.74 & \color{blue}{\bf 94.35} & 90.20& 80.92 & 91.29 & 90.19  &{\bf{93.54}} & { 93.31}\\
   Credit & NA &  NA & 95.31 & NA & NA & 94.83 & 95.24& 63.88 & 82.69 & 93.87  &{\bf{95.36}} & \color{blue}{\bf 95.65}\\
    
     \hline
     Average-AUC &81.55& 68.45& 91.01& 78.56& 74.92& 88.42& 88.56& 80.68&
       75.08& 88.51& \color{blue}{\bf{92.76}}& \bf{93.16}\\
Median rank &7.5&  11& 4.5& 9.5& 7.5& 4.5& 4& 8& 9.5& 5&  2.5& 3\\
     \bottomrule
    \end{tabular}%
      \end{adjustbox}
        \caption{Performance comparison with several leading anomaly detection baselines in the ``out-of-sample'' setting. We present the median AUC over $10$ runs. }
  \label{tab:realresults_oos}%
  \vskip -0.2 in
\end{table*}%

\begin{table}[]
  \begin{adjustbox}{width=1\columnwidth,center}
\begin{tabular}{lcccccccccccc}
\hline
Data set    & \multicolumn{1}{l}{CBLOF} & \multicolumn{1}{l}{COF} & \multicolumn{1}{l}{IForest} & \multicolumn{1}{l}{SOD} & \multicolumn{1}{l}{LSCP} & \multicolumn{1}{l}{HBOS} & \multicolumn{1}{l}{OC-SVM} & \multicolumn{1}{l}{DSVDD} & \multicolumn{1}{l}{RSR-AE} & $\ell_{2,1}$-AE & \multicolumn{1}{l}{PRAE-$\ell_0$} & \multicolumn{1}{l}{PRAE-$\ell_1$} \\ \hline
Thyroid     & 0.38                      & 0.11                    & \color{blue}{\bf{0.61} }                       & 0.33                    & 0.35                     & \bf{0.53 }                    & 0.48                       & 0.22                      & 0.15                       & 0.41            & 0.45                              & 0.43                              \\
Cardio      & 0.50                      & 0.22                    & 0.52                        & 0.33                    & 0.30                     & 0.47                     & 0.59                       & 0.30                      & 0.18                       & \bf{ 0.64}            & \color{blue}{\bf{0.65} }                             & 0.63                              \\
Ecoli       & 0.50                      & 0.67                    & 0.74                        & 0.50                    & \bf{ 0.80  }                   & 0.42                     & \color{blue}{\bf{ 0.81}   }                    & 0.70                      & 0.17                       & 0.09        &     0.76                              & 0.75                              \\
Lympho      & 0.50                      & 0.63                    & \bf{0.82  }                      & 0.62                    & 0.73                     & \color{blue}{\bf{ 0.92} }                    & 0.87                       & 0.28                      & 0.57                       & 0.82            & 0.62                              & 0.71                              \\
Pendigits   & 0.45                      & 0.08                    & 0.37                        & 0.10                    & 0.06                     & 0.35                     & 0.41                       & 0.18                      & 0.12                       & 0.32            & \bf{  0.63 }                             & \color{blue}{\bf{ 0.66  }  }                          \\
Yeast       & 0.20                      & 0.19                    & 0.34                        & 0.18                    & 0.23                     & 0.29                     & 0.34                       & 0.15                      & 0.33                       & 0.21            & \bf{ 0.39}                              & \color{blue}{\bf{  0.40 }  }                           \\
Musk        & 0.55                      & 0.11                    & 0.71                        & 0.35                    & 0.55                     & 0.84                     & 0.47                       & 0.70                      & 0.80                       & \color{blue}{\bf{ 0.98 } }          & 0.80                              & \bf{ 0.87 }                             \\
Mammography & 0.28                      & 0.18                    & 0.25                        & 0.17                    & 0.05                     & 0.18                     & 0.29                       & 0.29                      & 0.04                       & 0.28            & \color{blue}{\bf{ 0.30} }                             & \color{blue}{\bf{0.30}}                              \\
Shuttle     & 0.30                      & 0.14                    & 0.95                        & 0.20                    & 0.14                     & 0.96                     & \color{blue}{\bf{  0.96 } }                     & 0.51                      & 0.14                     & \bf{ 0.94 }           & 0.93                              & 0.92                              \\
MNIST-S     & 0.43                      & 0.47                    & 0.38                        & 0.31                    & 0.61                     & 0.36                     & 0.43                       & 0.28                      & 0.42                       & 0.44            &\bf{ 0.53 }                             & \color{blue}{\bf{ 0.65}   }                           \\ 
Fashion-MNIST     & 0.52                     & 0.27                   & 0.38                        & 0.24                    & 0.34                     & 0.22                     & 0.51                      & 0.29                      & 0.34                       & 0.55            &\color{blue}\bf{ 0.62 }                             & {\bf{ 0.59}   }                           \\ \hline
Average-F1  & 0.42                     & 0.28                  & 0.55                       & 0.31                   & 0.38                    & 0.50                    &          0.56                  &                 0.36          &         0.29                   & 0.52            & \bf{ 0.61 }                             & \bf{ 0.63 }                             \\
Median rank & 7                       & 10                       & 5                           & 10                       & 8                        & 6                        &                    4      &                 9          &            10                & 5               & 3                                 & 2                                 \\ \hline
\end{tabular}
    \end{adjustbox}
    \caption{Performance comparison with several leading anomaly detection baselines. We present the median maximal F1 score (where the maximum is computed for each run). }
      \label{tab:realresults_f1}
\end{table}

\begin{figure}[htb!]
  \vskip -0. in
    \centering 
        \includegraphics[width=0.44\textwidth]{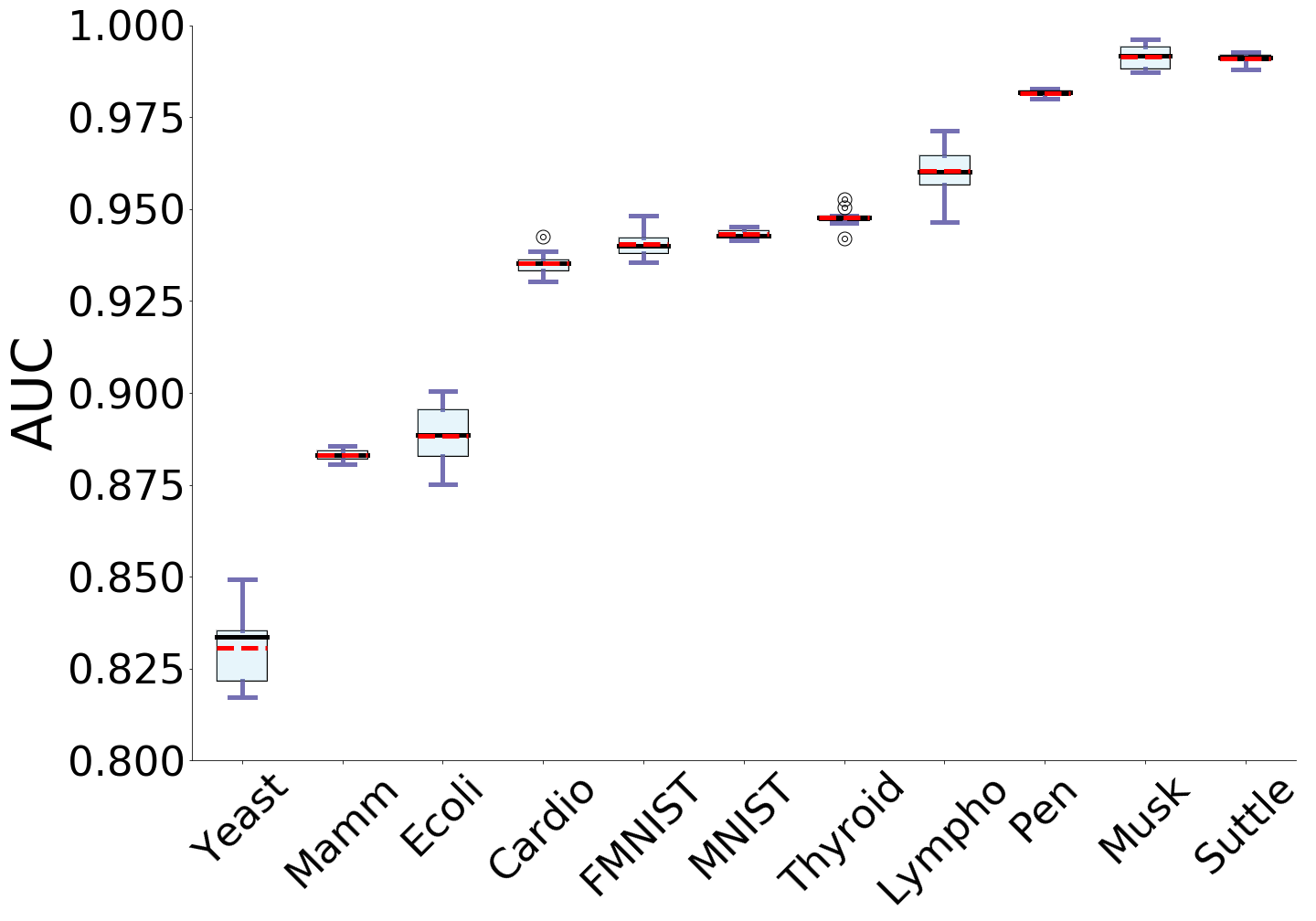}
    \includegraphics[width=0.44\textwidth]{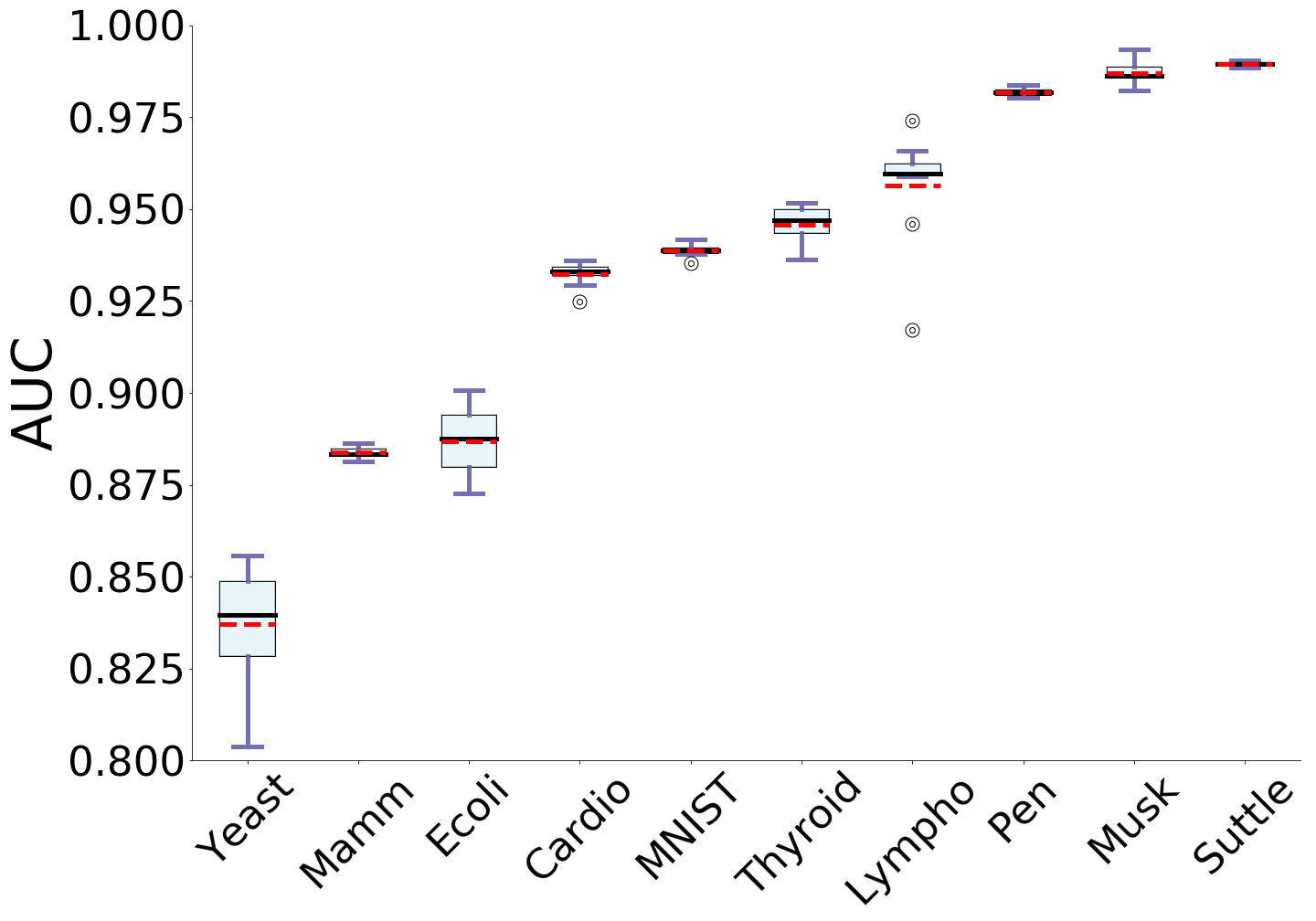}
  \vskip -0.1 in
    \caption{Box plots presenting the AUC of the proposed approach on several real datasets. The top and bottom plots represent PRAE-$\ell_0$ and PRAE-$\ell_1$ respectively. We use $10$ different random initializations on each dataset and compute the AUC after convergence.}
    \label{fig:box_plots}
      \vskip -0.1 in
\end{figure}

Next, to evaluate the stability of the proposed approach, we run the method with $10$ random initialization on each of the real datasets. In Figure \ref{fig:box_plots} we present box plots indicating the mean, median and $25/75$ percentiles of PRAE-$\ell_0$ and PRAE-$\ell_1$ (left and right panels). As evident from this figure, the interquartile range (IQR) of the proposed approach is relatively small for larger datasets (such as Shuttle, Pen, and MNIST-S). Finally, we evaluate the maximal F1 score attained by each method (when varying the anomaly score threshold). The median of this value across ten runs is reported in Table \ref{tab:realresults_f1}. 
\section{Datastes}
All real world datasets analyzed in this study are publicly available. Most datasets can be directly downloaded from \footnote{http://odds.cs.stonybrook.edu/}. MNIST and Fashion-MNIST can be easily obtained using python from ``keras'' package. The Purified populations of peripheral blood monocytes (PBMCs) is a Single-cell RNA sequencing (scRNA-seq) data. It was collected by \cite{zheng2017massively} and contains more that $90,000$ cells with $32,738$ genes. We randomly sample $6,000$ CD34 cells as out normal samples, and add $300$ cells from the remaining $8$ cell types. Since typically in scRNA-seq there is some level of false annotations, we design this example to evaluate the capabilities of PRAE in curating a set of CD34 from contamination by other cell types. 
\section{Baselines and Hyperparameters}
In the following section, we describe all baselines and hyperparameters used for the evaluation of anomaly detection using real and synthetic data. We train our proposed AE with an encoder-decoder pair with five hidden layers each of size $10$; the hidden dimension is $1$ (this might not be optimal but worked well across most datasets). For the high dimensional datasets (PBMC, Fashion-MNIST) the hidden layers are of size $100$. We use the heuristic proposed in Section 6.2 
to tune the regularization parameter to $\lambda=1<ME$. We use Adam optimizer with a learning rate of $N\cdot10^{-6}$ where $N$ is the number of samples in the dataset. \\
{\bf Clustering-Based Local Outlier Factor (CBLOF) \cite{cblof} } is a proximity-based method that relies on clustering to define an anomaly score for each sample.\\
{\bf Angle-Based Outlier Detection (ABOD) \cite{abod}} uses vectors defined between pairs of points to identify outliers by comparing the angles between the different vectors.\\
{\bf Connectivity-Based Outlier Factor(COF) \cite{COF}} relies on proximity between samples to identify outliers.\\
{\bf Isolation Forest (IForest) \cite{Iforest}} seeks the minimum number of splits required to isolate each sample. Then, an ensemble of trees are used to define an anomaly score.\\
{\bf Subspace Outlier Detection (SOD) \cite{SOD}} identifies anomalies as samples that deviate significantly from the subspace spanned by a subspace defined based on neighbors of point.\\
{\bf Locally Selective Combination of Parallel Outlier Ensembles (LSCP) \cite{lscp}} is an ensemble method that uses multiple local subspaces to identify outliers.\\
{\bf Histogram-based Outlier Score (HBOS) \cite{HBOS}} is a probabilistic approach that creates histograms to identify anomalies as samples with low density.\\
{\bf One-Class Support Vector Machines (OC-SVM) \cite{scholkopf2001estimating}} uses support vectors to identify the margins of the normal part of the data. Here, we use a Gaussian kernel to capture the non-linearity of this method.\\
{\bf Deep One-Class Classification (Deep-SVDD) \cite{ruff2018deep}} extends OC-SVM by introducing a NN to model the decision boundary of the normal data. Here, we evaluated several architectures for DSVDD and selected the one that leads to the smallest validation error. The method is optimized with Adam and the number of epochs in $100$.  \\
{\bf Robust deep autoencoders ($\ell_{2,1}-AE$) \cite{zhou2017anomaly}} uses an $\ell_{2,1}$ to regularize an AE that attempts to reconstruct the data while removing outliers. The approach is explained in the main text. We use an AE with the same architecture as the proposed approach in all examples.\\
{\bf Ensemble of Autoencoders (RandNet) \cite{chen2017outlier}} In the absence of an implementation of the method, we report the results presented by the authors. The method relies on an aggregated of an ensemble of AEs. The AUC of the method for $6$ out of the to $13$ analyzed datasets are reported in Table \ref{table:randnet}. Our method outperforms Randnet and the majority of these datasets.

\begin{table}[!ht]
    \centering
    \begin{tabular}{|c|c|c|}
    \hline
        dataset & RandNet & Rank\\ \hline
        Thyroid & 90.42 & 6 \\ \hline
        Cardio & 92.87 & 3 \\ \hline
        Ecoli & 85.42  & 9 \\ \hline
        Lympho & 99.06 & 2 \\ \hline
        Pendigits & 93.44 & 5\\ \hline
        Yeast & 82.95 & 3\\ \hline
    \end{tabular}\caption{AUC results of RandNet as reported by the authors along with the rank compared to all existing baselines. These were omitted from the table in the main text since the datasets only partially overlap.}\label{table:randnet}
\end{table}

\section{Sensitivity to Hyperparameters}
\begin{figure}[htb!]
  \vskip -0. in
    \centering 
        \includegraphics[width=0.44\textwidth]{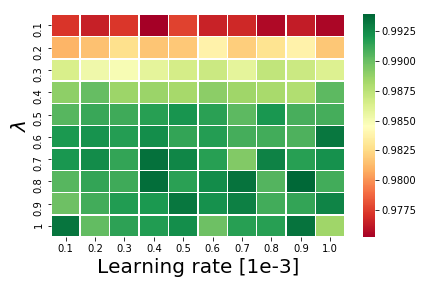}
    \includegraphics[width=0.44\textwidth]{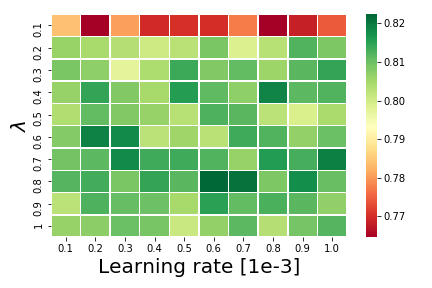}
  \vskip -0.1 in
    \caption{Heatmap presenting the AUC of the proposed approach on for several values of $\lambda$ and the learning rate. The left plot indicates performance on `Musk' data, while the right plot is based on the `Yeast' data.}
    \label{fig:sensitiv}
      \vskip -0 in
\end{figure}

In the following experiment, we evaluate the sensitivity of the method to hyperparameters.
Towards this goal, we run PRAE on the `Yeast' and `Musk' datasets for various values of $\lambda$ and different learning rates. These datasets where arbitrarily chosen from the datasets in Table 2.
For each value of the parameters we repeat the experiment ten times and report the median AUC. 
In Figure \ref{fig:sensitiv} we present heatmaps with the median AUC values for each set of parameters used in the evaluation. 
As demonstrated by this figure, PRAE is relatively stable to both of these hyperparametrs, and the overall AUC varies by less than $10\%$ across all evaluated values of $\lambda$ and the learning rate. 
As expected, if the value of $\lambda$ is small, the performance of the model deteriorates. 
This could happen if the model removes too many inliers from the objective at an early stage of training, thus "hurting" the ability to distinguish between inliers and outliers. 

Another hint for the stability of the algorithm to the hyperparameter $\lambda$ is the fact that we used the same $\lambda$ for all experiments in Table 2
and did not choose a different value for each experiment.

\section{Ablation Study}
In this section, we perform an ablation study to evaluate the influence of each element of our algorithm. We focus on the `Yeast' and `Musk' datasets, and compare PRAE-$\ell_0$, and PRAE-$\ell_1$, to the following variants:
\begin{itemize}
    \item {\bf AE}- a standard AE with no regularization, anomalies are identified using the reconstruction error.
    \item {\bf DRAE-$\ell_1$}- a variant of PRAE, but with deterministic gate values, and a standard $\ell_1$ regularizer.
    \item {\bf PRAE $(\lambda=0)$}- a variant of PRAE, but with a regularization term$=0$.
\end{itemize}
As demonstration in Table \ref{tab:ablation} the proposed probabilistic regularization leads to improved identification of outliers compared with all variants of the method. This empirical results, suggests that removing outliers throughout training with the stochastic gates can lead to more reliable identification of outliers using AEs.

\begin{table}[!ht]
    \centering
    \begin{tabular}{|c|c|c|c|c|c|}
    \hline
        dataset & AE & DRAE-$\ell_1$ & PRAE $(\lambda=0)$  & PRAE-$\ell_0$ & PRAE-$\ell_1$ \\ \hline
        Yeast & 74.05 & 77.26 & 77.18 & 83.37 & 83.95 \\ \hline
        Musk & 95.61 & 96.81 & 88.63 & 99.17 & 98.61 \\ \hline
    \end{tabular}\label{tab:ablation}
    \caption{Ablation study. Comparing the proposed schemes PRAE-$\ell_0$, and PRAE-$\ell_1$, to three other variants. We use the same architecture, optimizer and compare to AE, DRAE-$\ell_1$ and PRAE $(\lambda=0)$ all explained above.  }
\end{table}
\section{Synthetic Swiss Roll}

\begin{figure}[htb!]
  \vskip -0. in
    \centering 
        \includegraphics[width=0.3\textwidth]{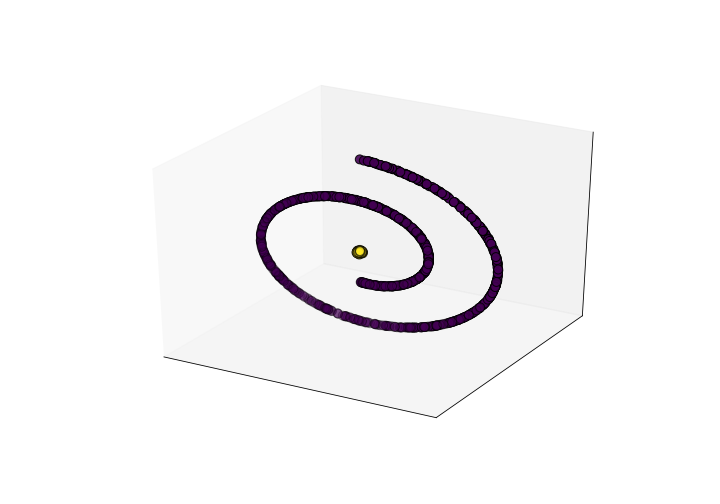}
    \includegraphics[width=0.3\textwidth]{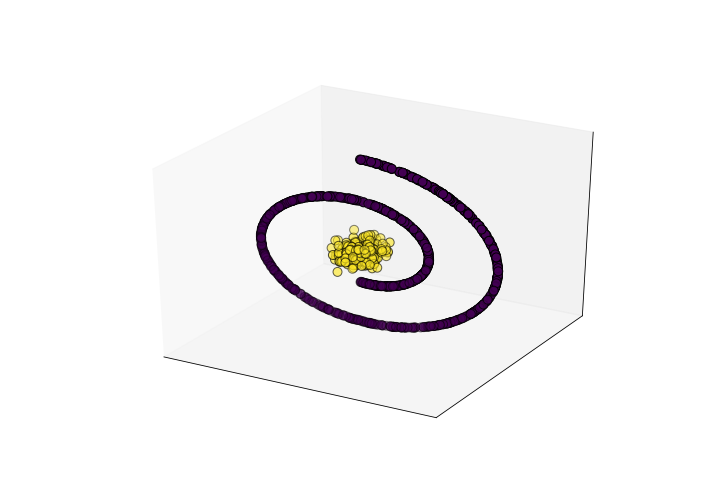}
        \includegraphics[width=0.3\textwidth]{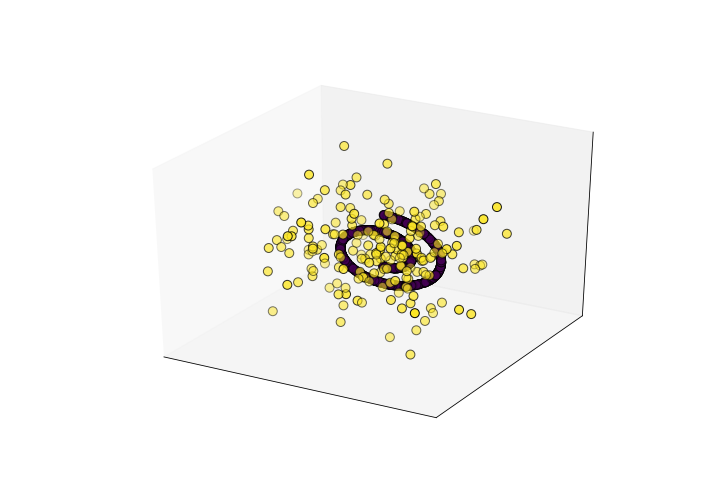}
  \vskip -0.1 in
    \caption{Examples of the synthetic narrow Swiss roll (blue) with Gaussian outliers (yellow), generated using different values of $\sigma^2\in{0.1,1,10}$.}
    \label{fig:swiss}
      \vskip -0.1 in
\end{figure}
In the synthetic example described in Section 6.3
, we consider a ``narrow swiss-roll'', with $1000$ points uniformly sampled from $ [ 3\pi/2, 9\pi/2] \times [0,0.1]$, and embedded into $\mathbb{R}^3$ using $(t,h)\rightarrow (t\cos(t),h,t\sin(t))$. 
Then, we generate additional $200$ ``outliers'' sampled from $N(0,\sigma^2 I_3)$, where $I_3\in \RR^{3\times 3}$ is the identity matrix. In Figure \ref{fig:swiss} we present examples of Swiss rolls with anomalies generated using different values of $\sigma$. In this example, we use a NN with hidden layers of size $512,256,128,64,32$, and the latent space has two neurons. Since the energy of the data varies substantially across samples, we used a normalized reconstruction loss for training all AE-based methods. Specifically, we normalize the reconstruction error of each sample by the $\ell_2$ norm of the sample.

\section{MNIST-S and Fashion MNIST}
In this section, we describe the more experiments performed using MNIST-S and Fashion MNIST.
\subsection{Small MNIST Dataset (MNIST-S) }
\begin{figure*}[tb!]
    \centering  
    \includegraphics[width=0.25\textwidth]{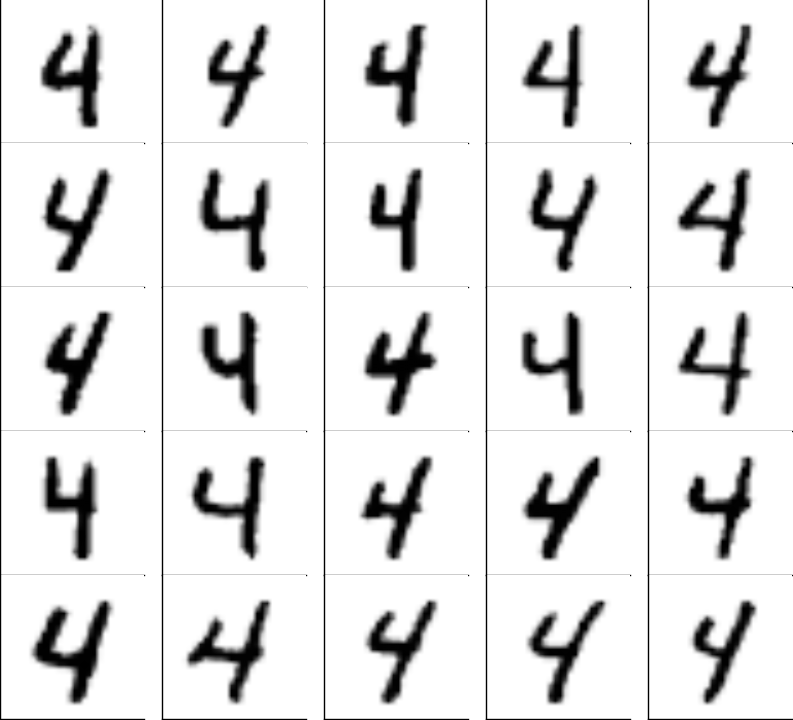} \hfill
        \includegraphics[width=0.25\textwidth]{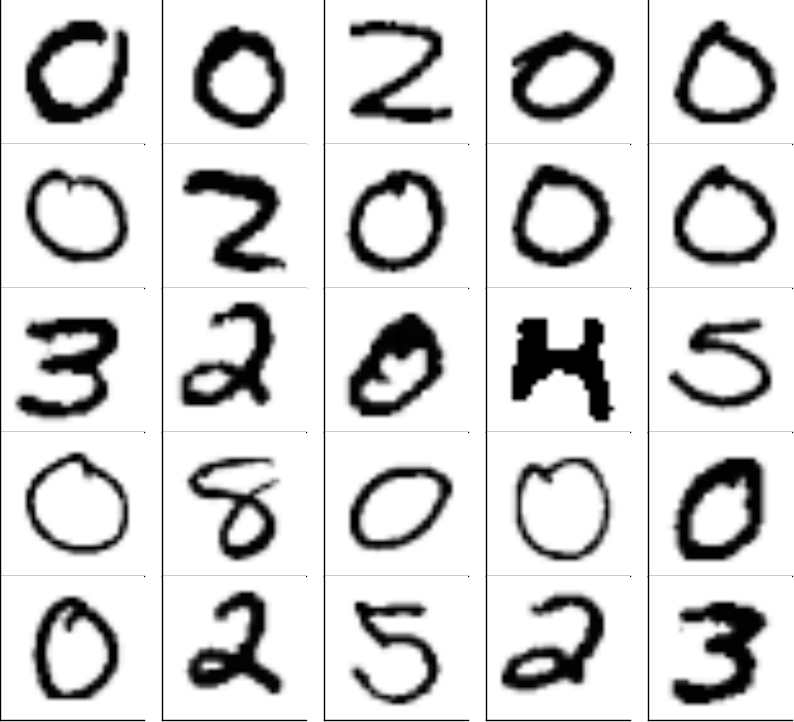}\hfill
            \includegraphics[width=0.32\textwidth]{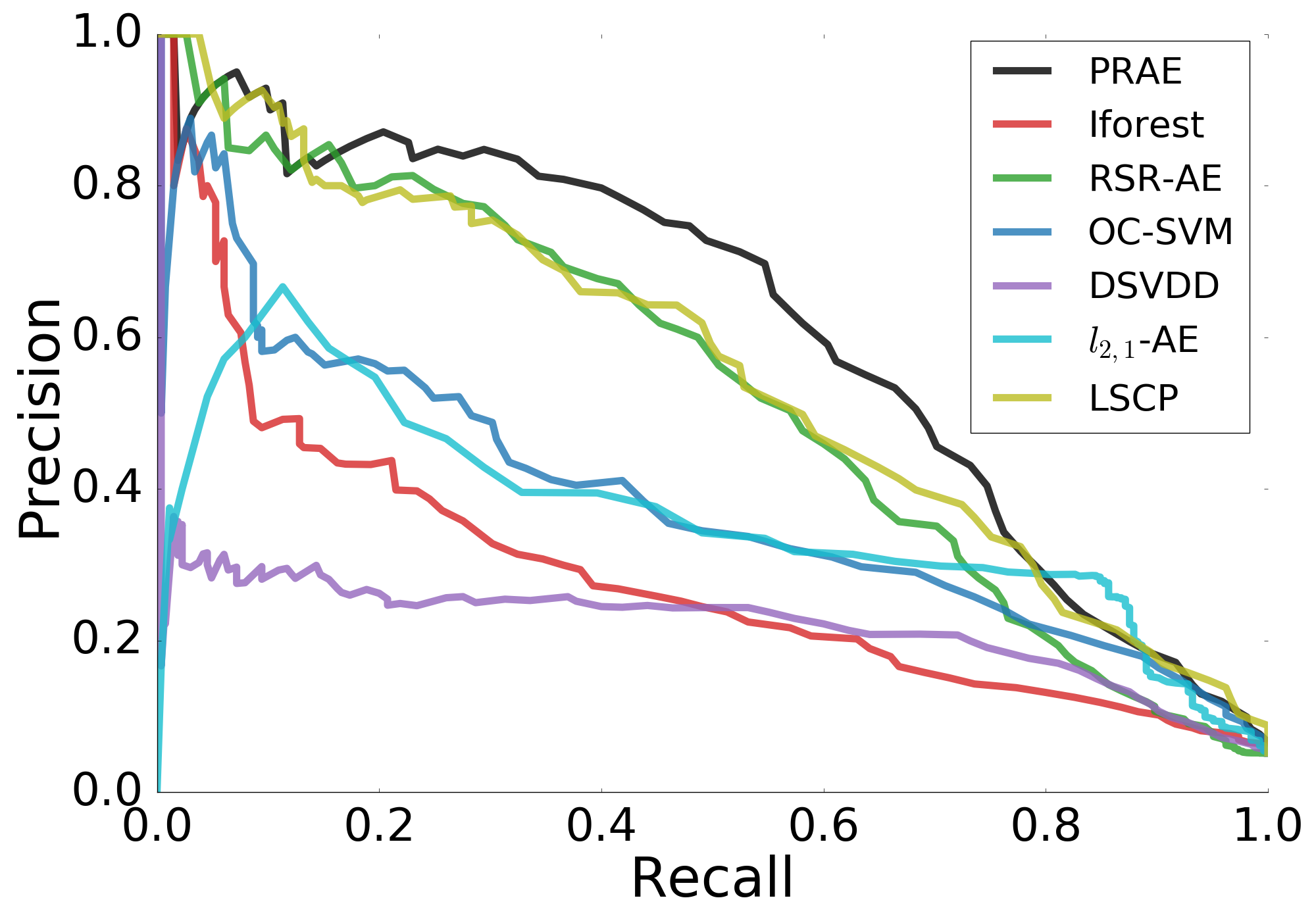}
    \caption{$25$ most inlaying/outlying (left/center) MNIST-S images as identified by PRAE. Right- precision vs. recall curves for leading baselines on the MNIST-S data. }
    \label{fig:mnist}
\end{figure*}

MNIST-S was proposed in \cite{zhou2017anomaly} for anomaly detection. 
To construct MNIST-S, we mix $4859$ nominal instances of the digit '4' with $265$ anomalies randomly sampled from all other digits. 
Following \cite{zhou2017anomaly}, we use a linear AE with one hidden layer of size $24$. 
Evaluation of the AUC of our method compared to all baselines appears in Table 3. 
This example appears to be especially challenging for density/distance-based baselines; we believe that this is due to the relatively high dimensionality of this data. 
In Figure \ref{fig:mnist}, we present the $25$ most \textit{inlaying} images (left panel), and the $25$ most \textit{outlying} images (center panel) as identified PRAE-$\ell_0$. 
The identified inliers share a standard ``simple'' structure of the digit '4'. 
On the other hand, most identified outliers are indeed of different digits, except for one example, which is somewhat of an unusual instance of the digit '4'. 

\begin{figure}[h!]
    \centering  
      \begin{tabular}{c c}
        \textbf{\underline{SOD-Normal}} & \textbf{\underline{SOD-Anomaly}} \\
        \includegraphics[width=0.4\textwidth]{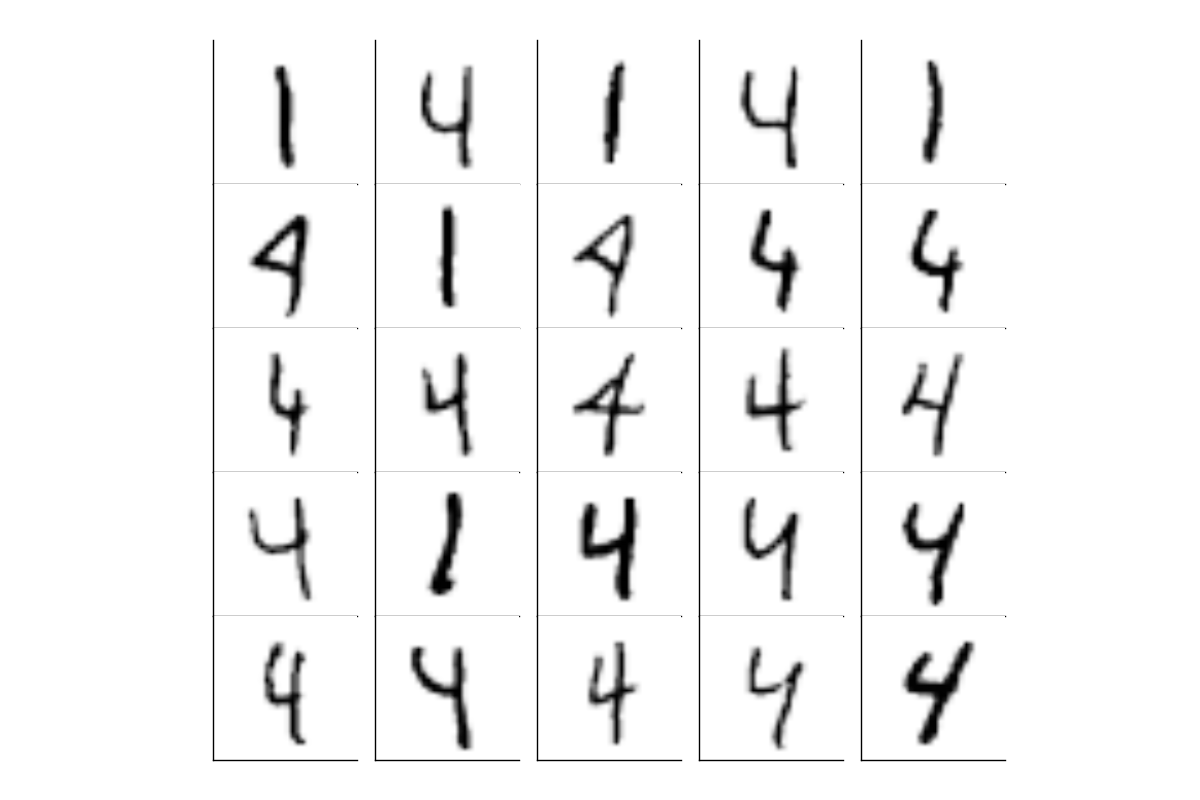}  & \includegraphics[width=0.4\textwidth]{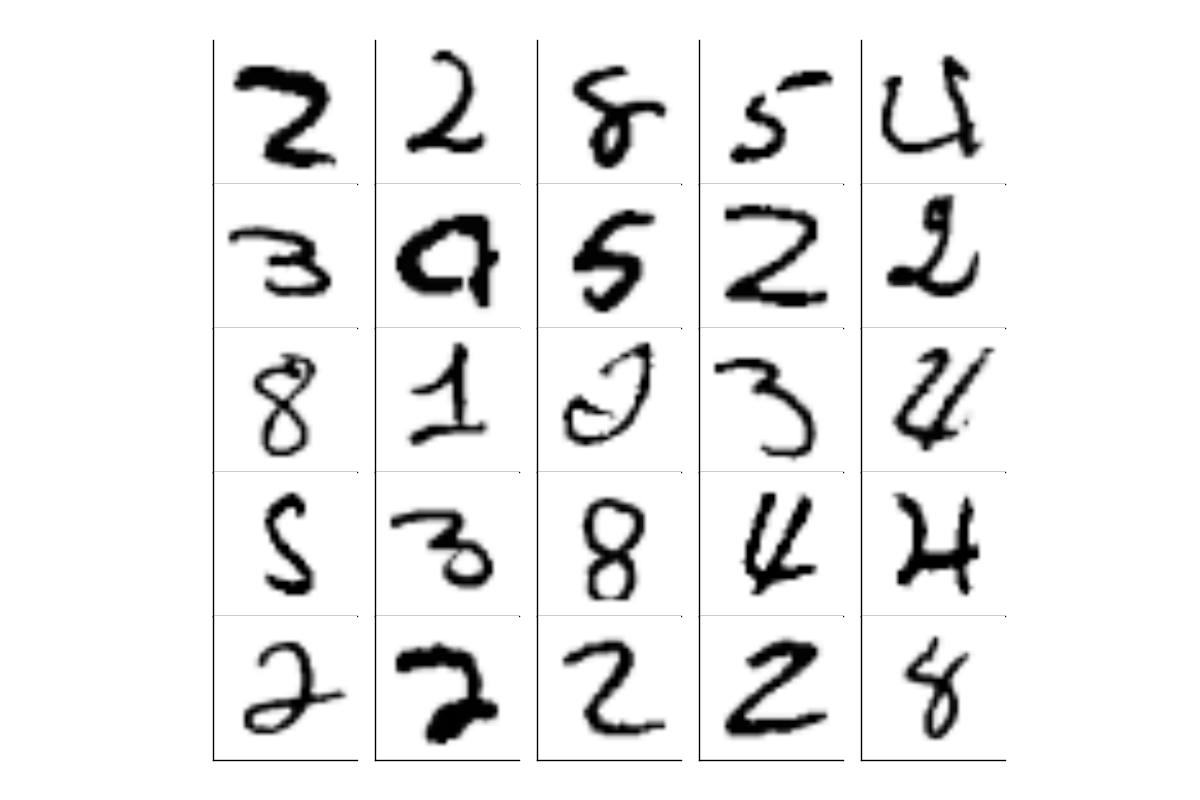} 
    \end{tabular}
          \begin{tabular}{c c}
        \textbf{\underline{LSCP-Normal}} & \textbf{\underline{LSCP-Anomaly}} \\
        \includegraphics[width=0.4\textwidth]{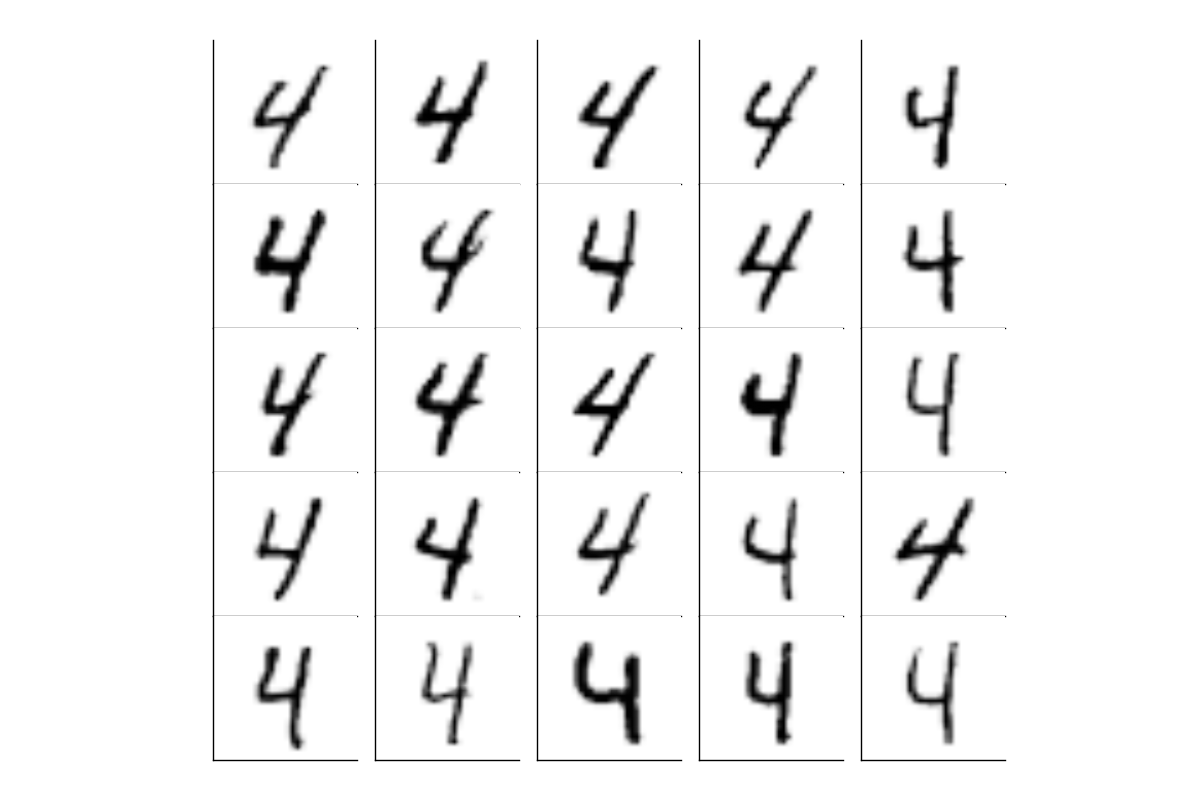}   &   \includegraphics[width=0.4\textwidth]{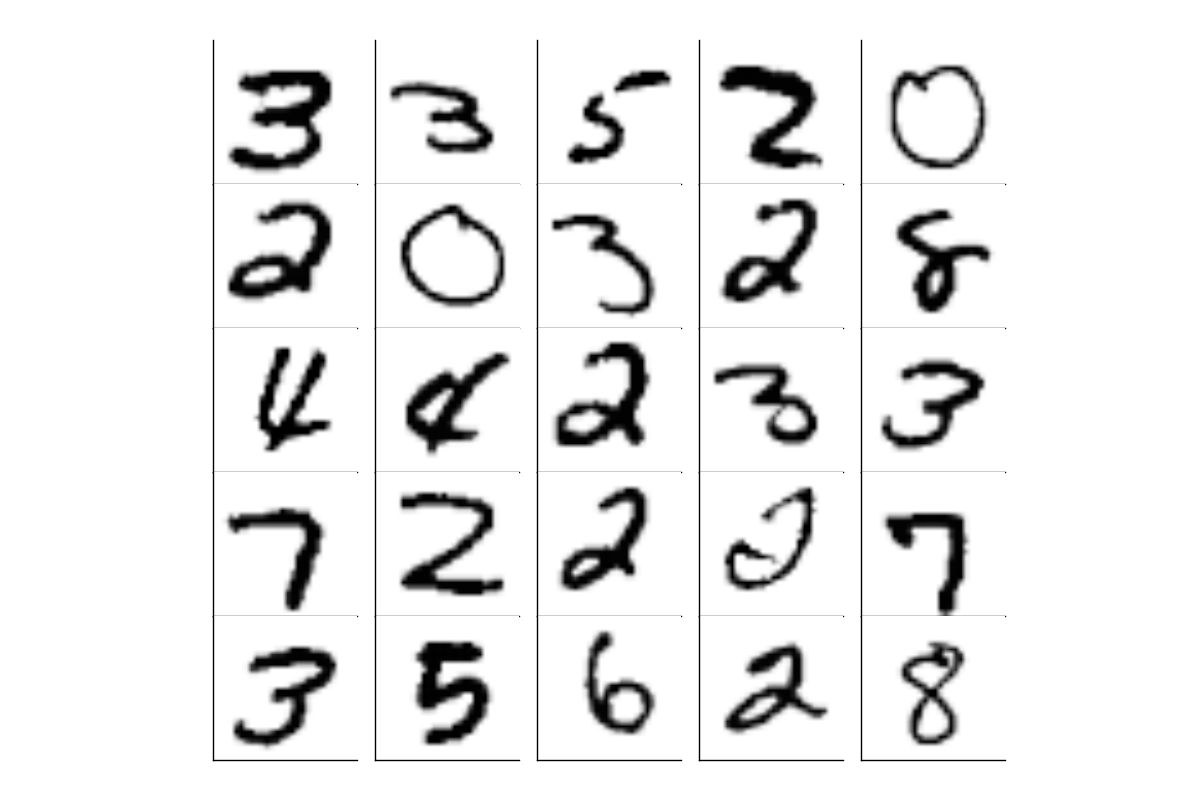}\
    \end{tabular}
          \begin{tabular}{c c}
        \textbf{\underline{HBOS-Normal}} & \textbf{\underline{HBOS-Anomaly}} \\
        \includegraphics[width=0.4\textwidth]{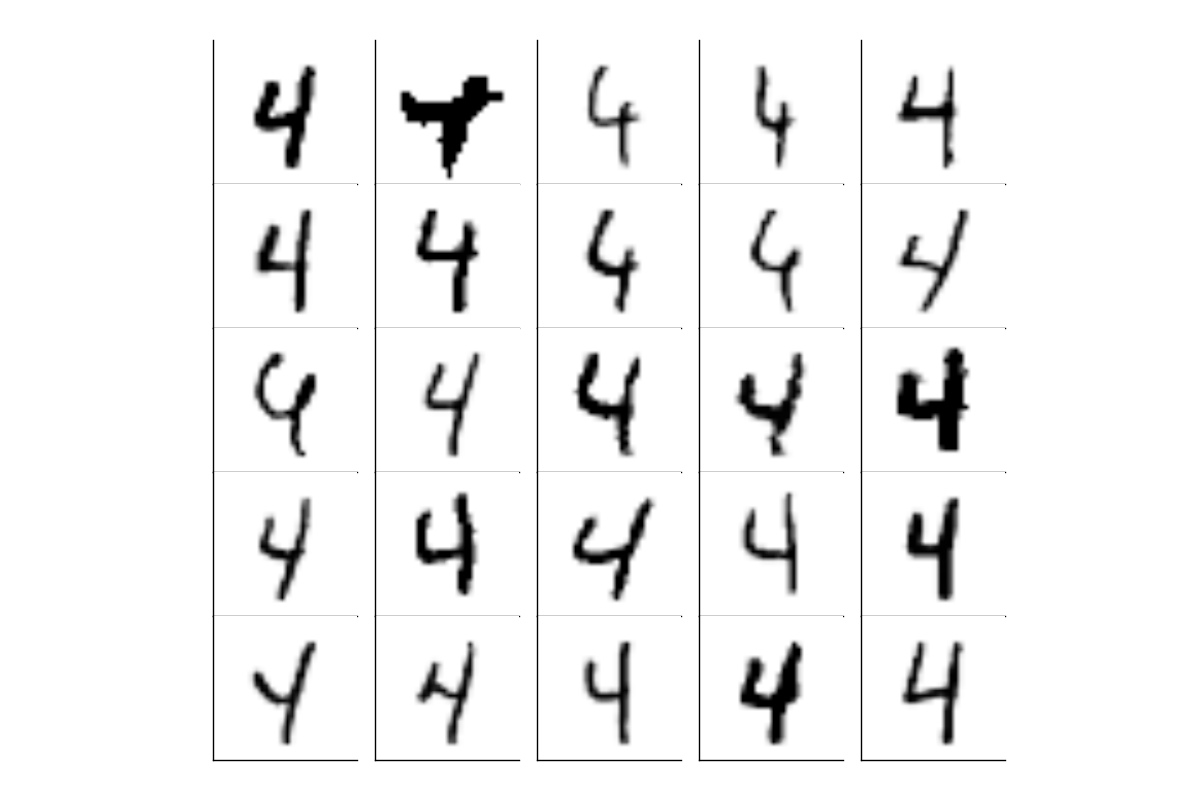}  & \includegraphics[width=0.4\textwidth]{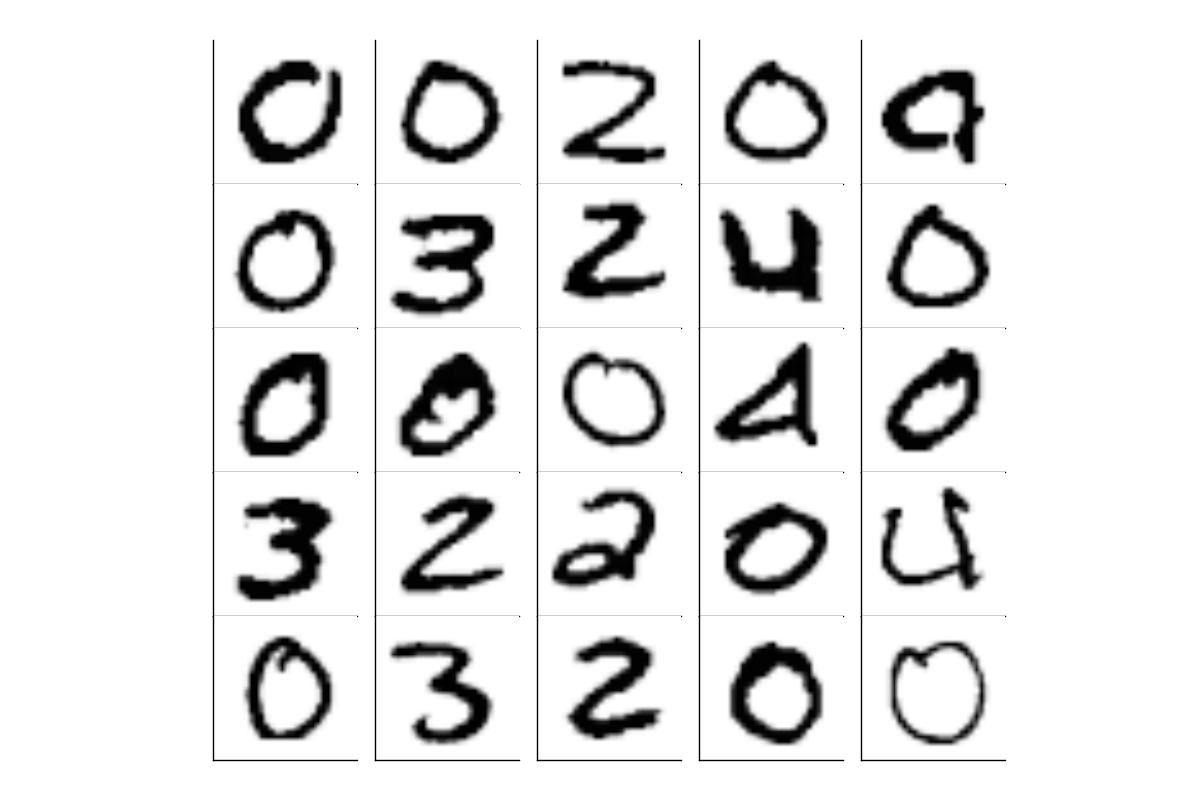} 
    \end{tabular}
          \begin{tabular}{c c}
        \textbf{\underline{ABOD-Normal}} & \textbf{\underline{ABOD-Anomaly}} \\
        \includegraphics[width=0.4\textwidth]{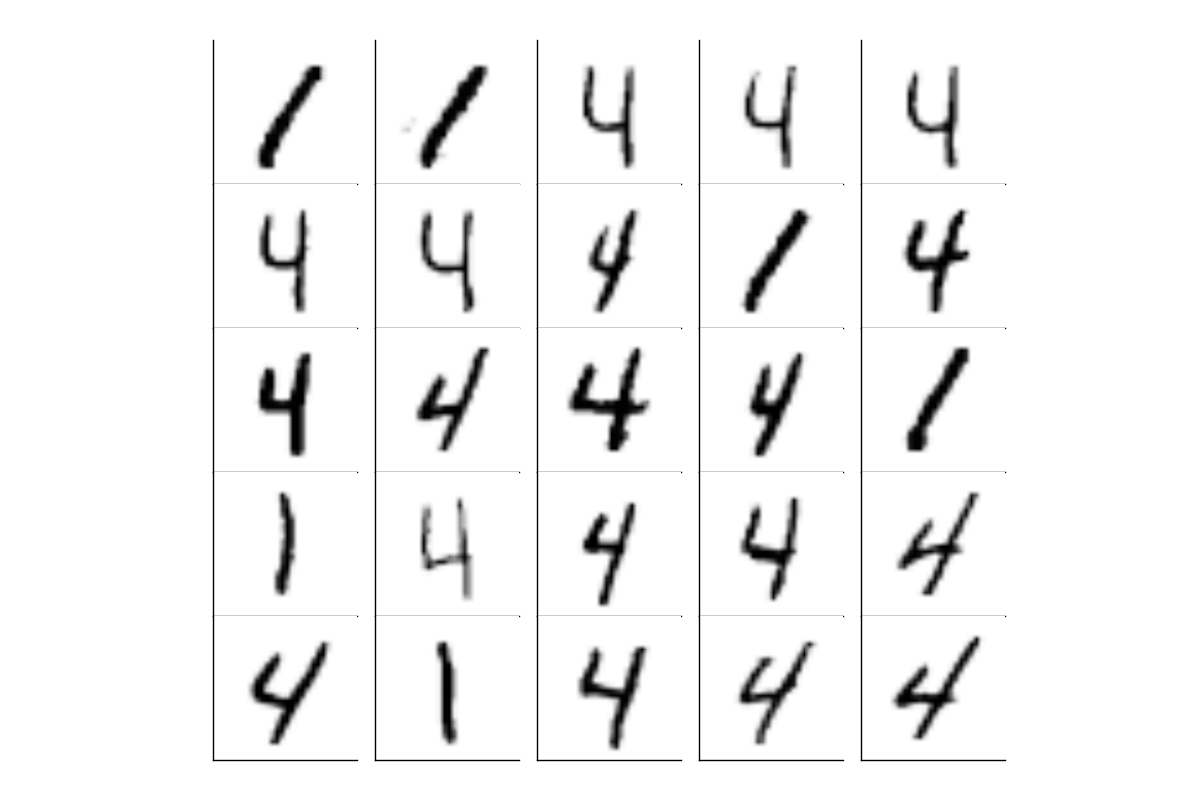}  & \includegraphics[width=0.4\textwidth]{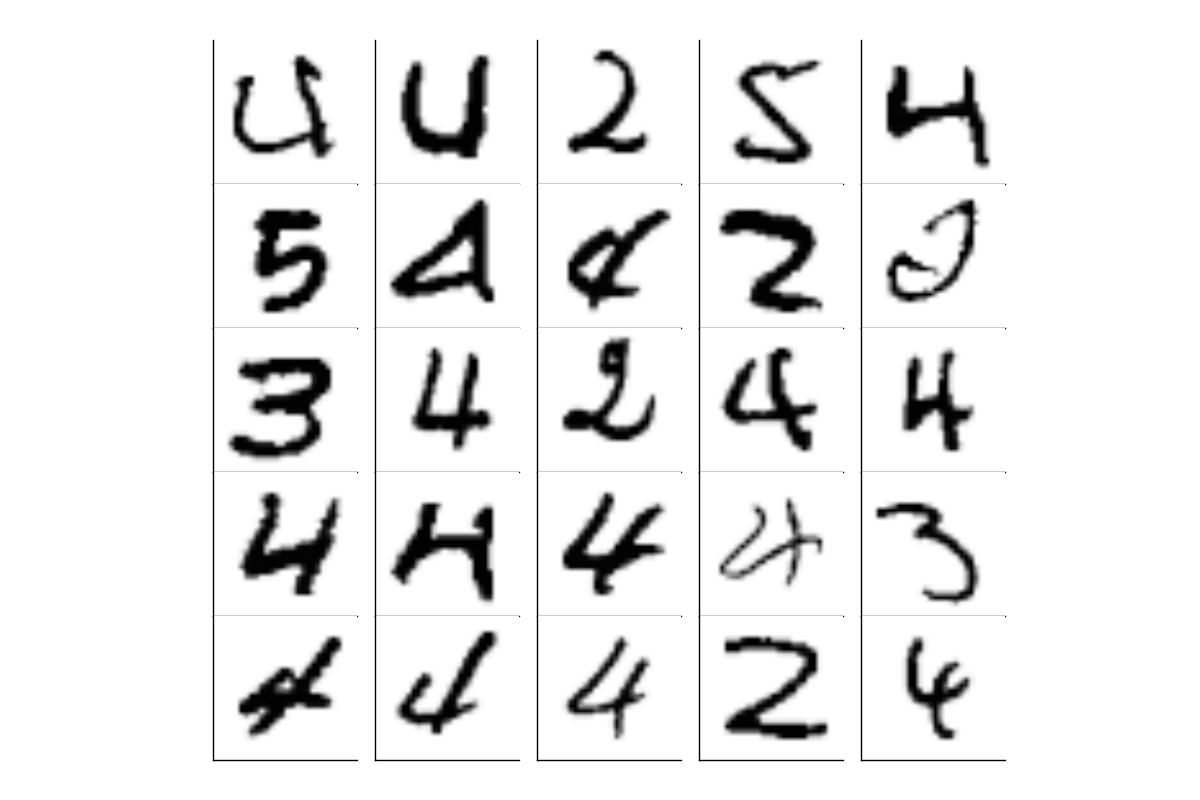} 
    \end{tabular}

    \caption{$25$ most inlaying/outlying (left/right) MNIST-S images as identified by different baseline algorithms. }
    \label{fig:mnist-baselines1}
\end{figure}

\begin{figure}[h!]
  \vskip -0.5 in
    \centering  
      \begin{tabular}{c c}
        \textbf{\underline{OC-SVM-Normal}} & \textbf{\underline{OC-SVM-Anomaly}} \\
        \includegraphics[width=0.4\textwidth]{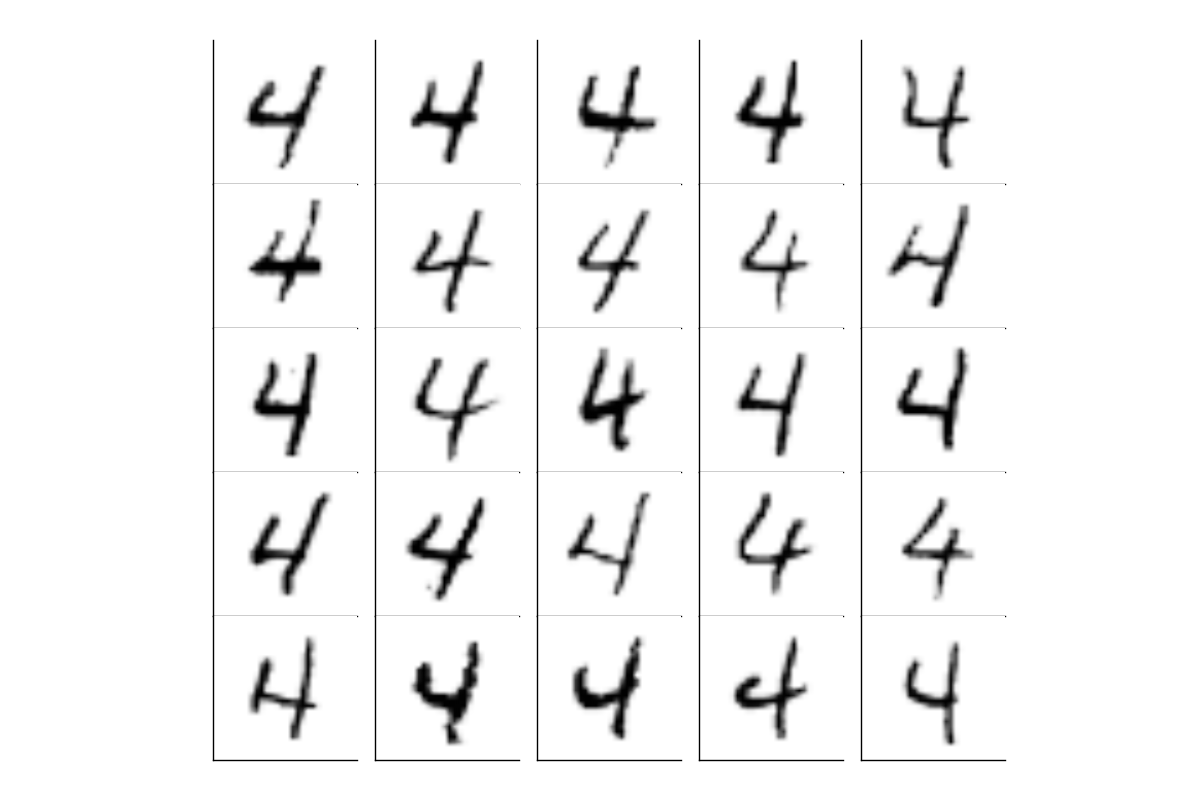}  & \includegraphics[width=0.4\textwidth]{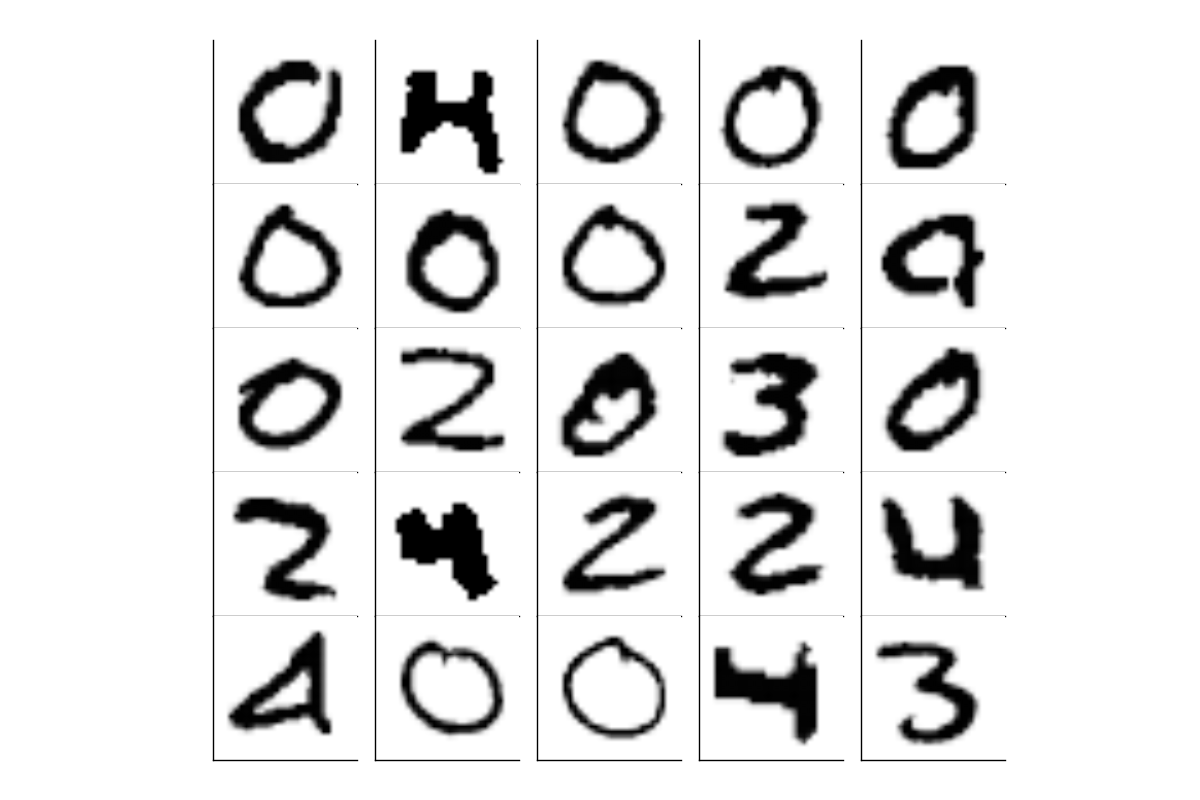} 
    \end{tabular}
          \begin{tabular}{c c}
        \textbf{\underline{DSVDD-Normal}} & \textbf{\underline{DSVDD-Anomaly}} \\
        \includegraphics[width=0.4\textwidth]{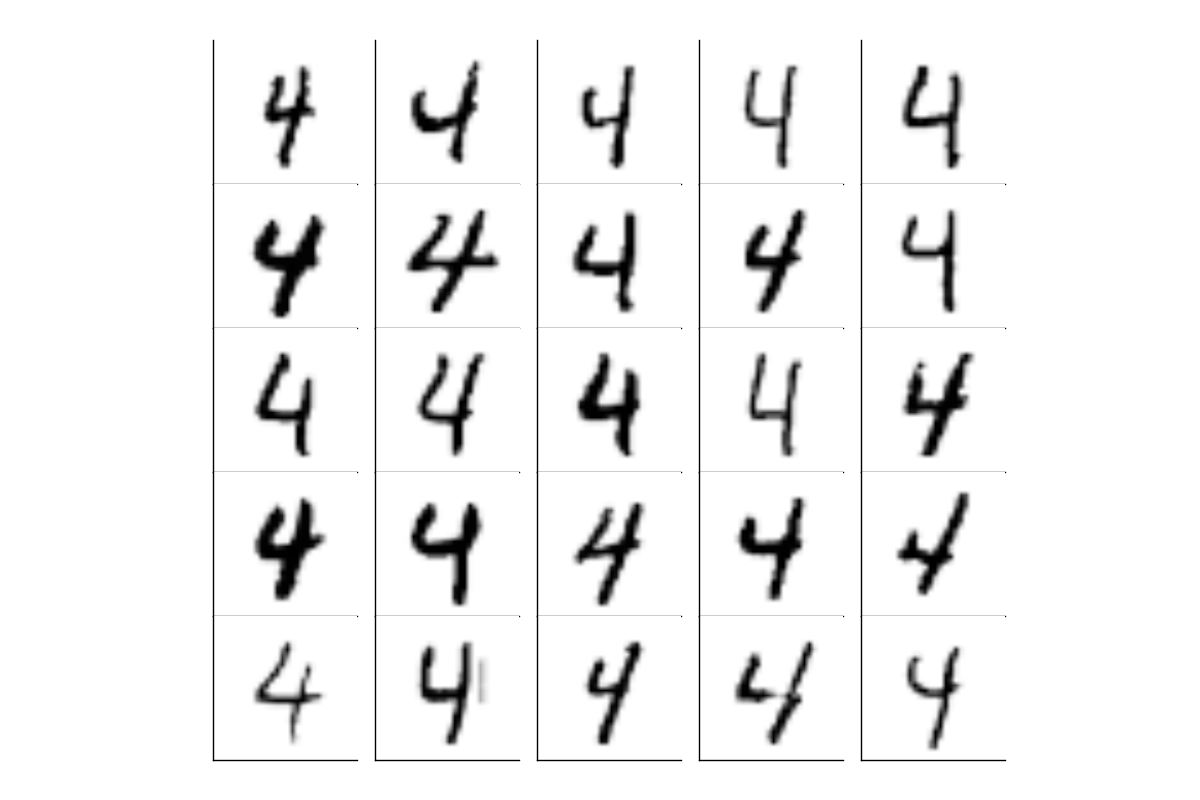}   &   \includegraphics[width=0.4\textwidth]{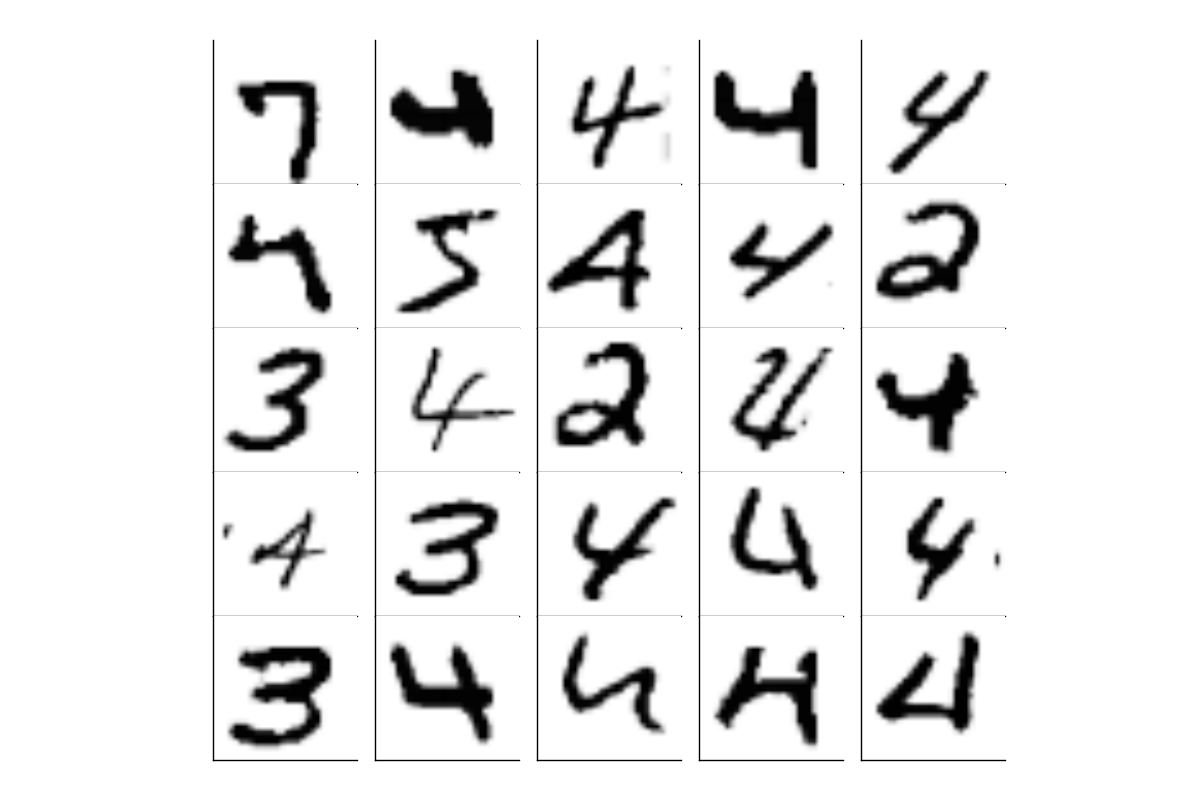}\
    \end{tabular}
          \begin{tabular}{c c}
        \textbf{\underline{$\ell_{2,1}$-AE-Normal}} & \textbf{\underline{$\ell_{2,1}$-AE-Anomaly}} \\
        \includegraphics[width=0.4\textwidth]{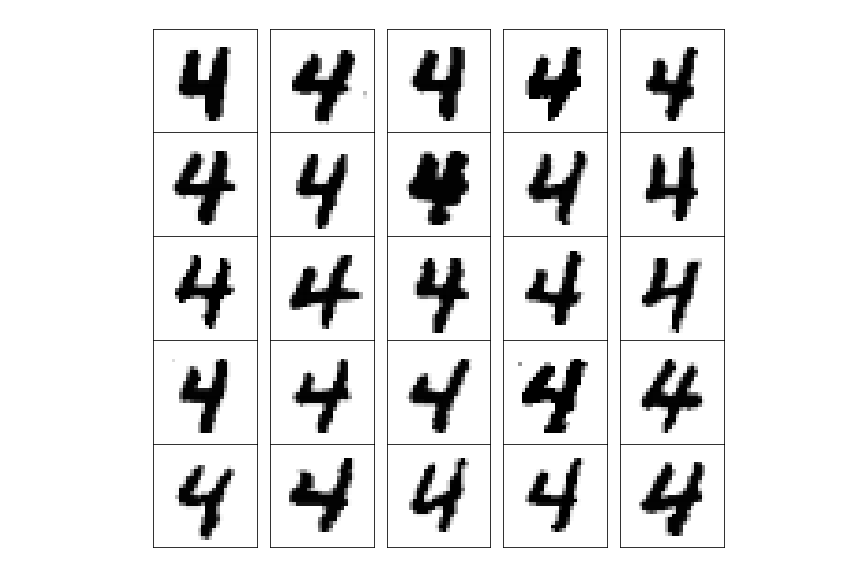}  & \includegraphics[width=0.4\textwidth]{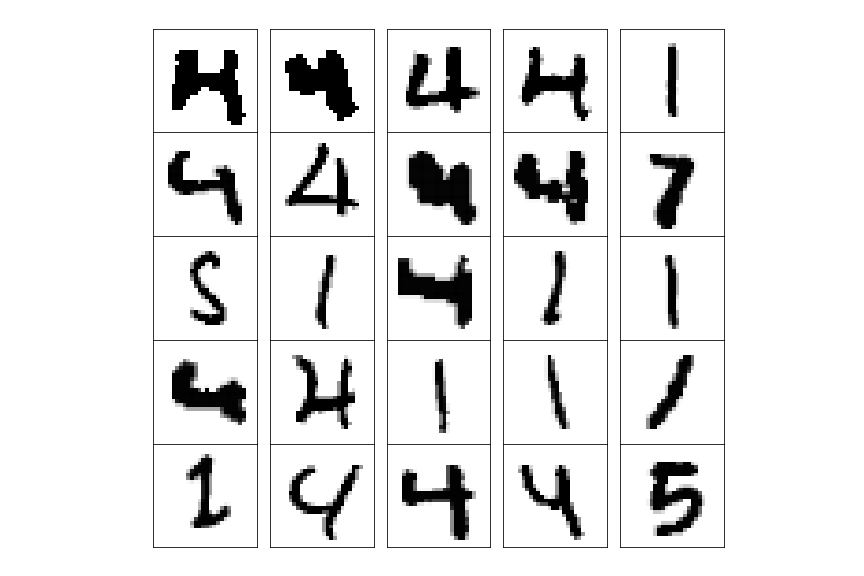} 
    \end{tabular}
          \begin{tabular}{c c}
        \textbf{\underline{RSR-AE-Normal}} & \textbf{\underline{RSR-AE-Anomaly}} \\
        \includegraphics[width=0.4\textwidth]{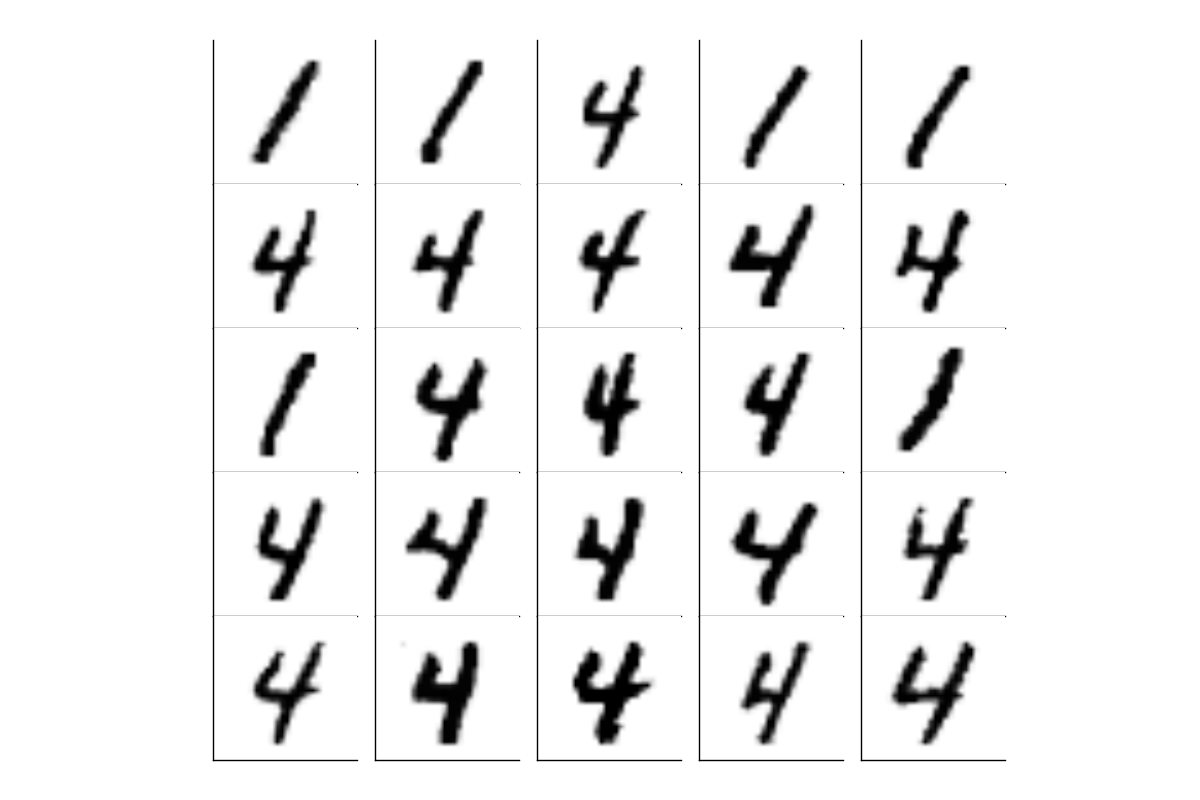}  & \includegraphics[width=0.4\textwidth]{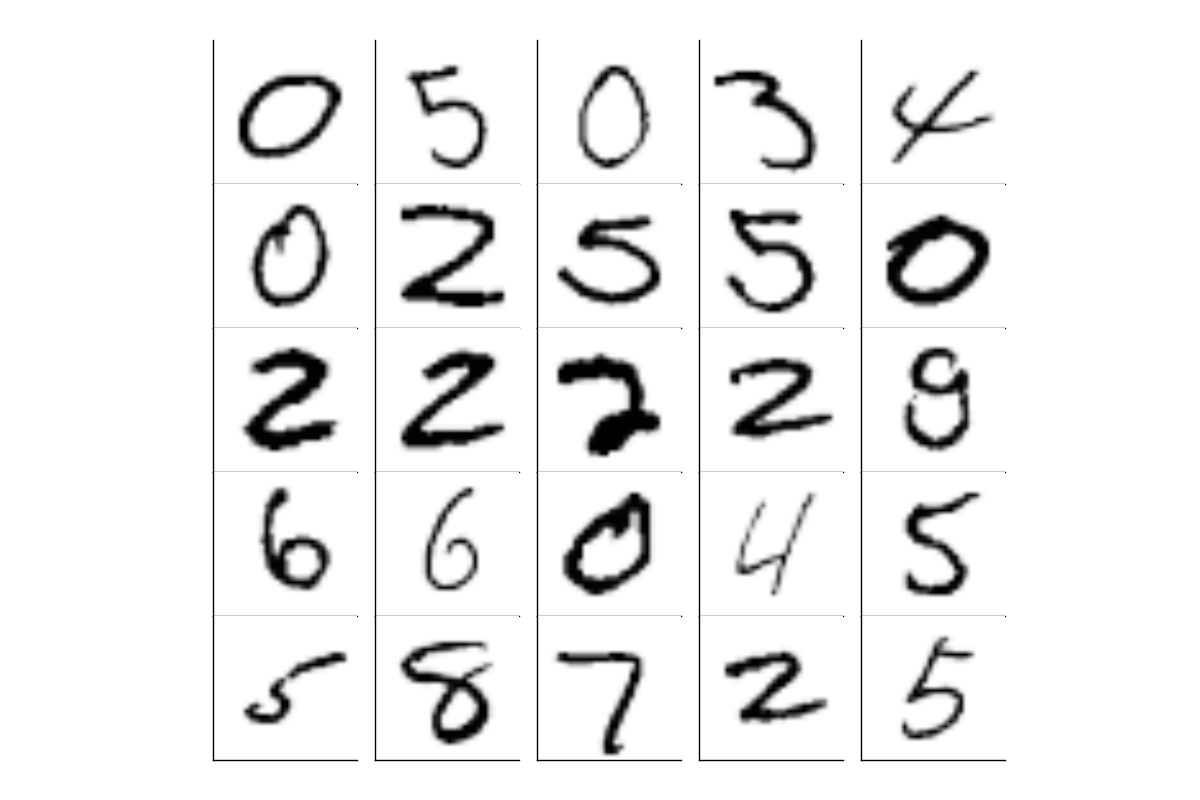} 
    \end{tabular}
              \begin{tabular}{c c}
        \textbf{\underline{IForest-Normal}} & \textbf{\underline{IForest-Anomaly}} \\
        \includegraphics[width=0.4\textwidth]{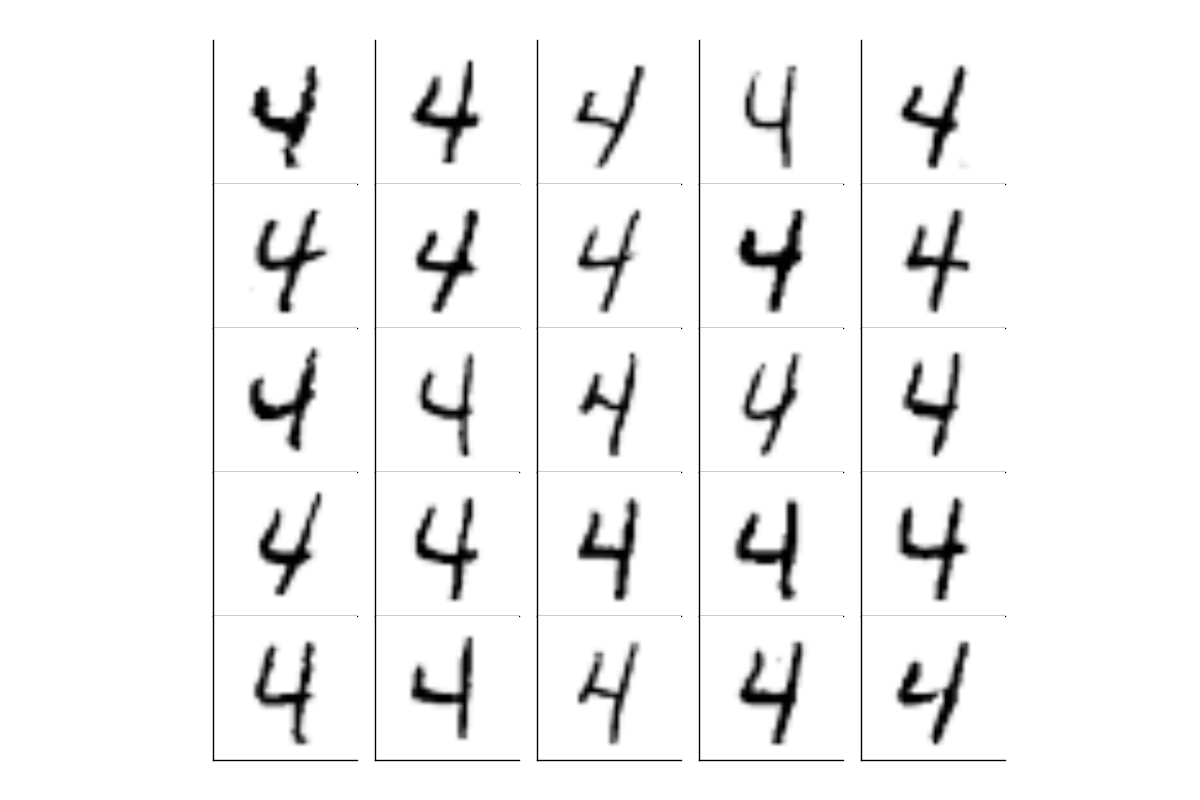}  & \includegraphics[width=0.4\textwidth]{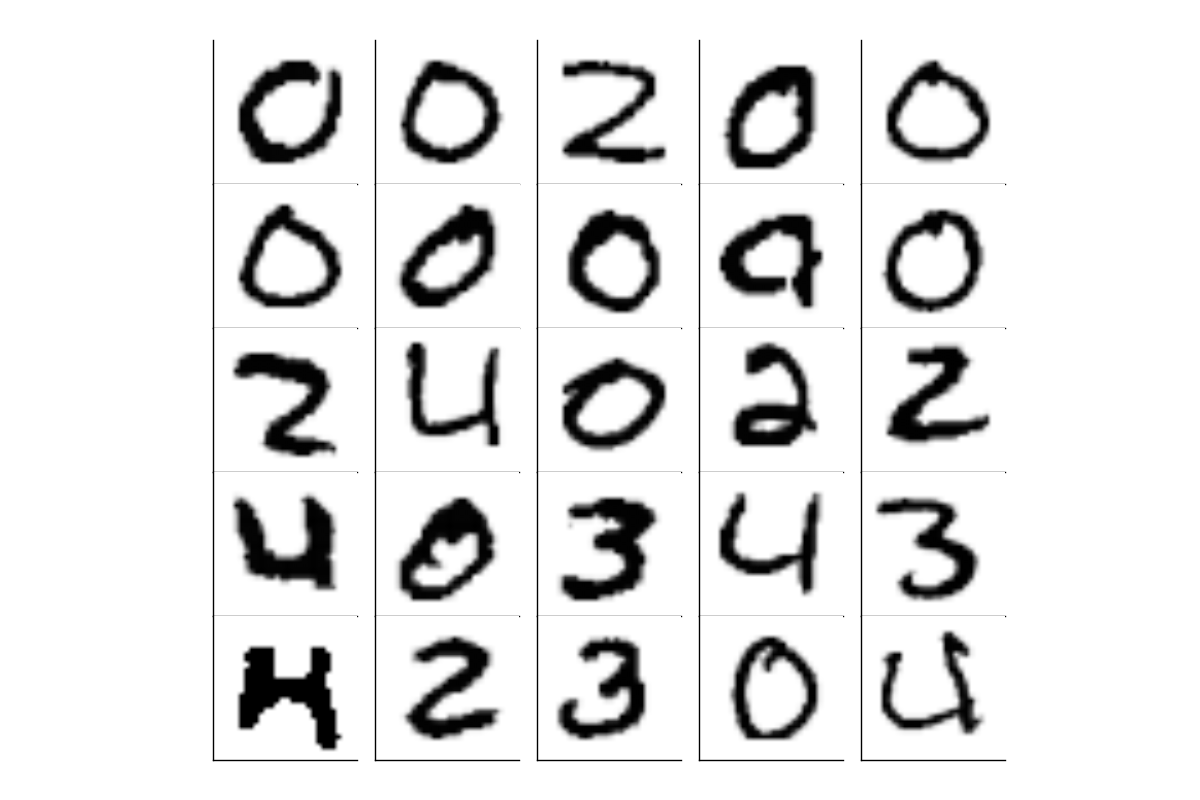} 
    \end{tabular}

    \caption{$25$ most inlaying/outlying (left/right) MNIST-S images as identified by different baselines}
    \label{fig:mnist-baselines2}
\end{figure}

To further assess the performance of the different baselines on MNIST-S we present the $25$ most  \textit{inlaying} images and the  $25$ most \textit{outlying} images as identified by the different baselines. These results are presented in Figures \ref{fig:mnist-baselines1} and \ref{fig:mnist-baselines2}.

Next, we repeat the MNIST-S experiment using different classes as the representatives for the normal samples. Specifically, we mix $4859$ nominal instances of the digit $C\in \{0,1,..., 9\}$ with $265$ anomalies randomly sampled from all other digits. We compare the AUC of PRAE to two leading baselines across all classes. As indicated in Table \ref{table:mist_all}, the proposed approach leads to more accurate outlier identification compared to the leading baselines across most of the classes. 

\begin{table}[!ht]
    \centering
    \begin{tabular}{|l|l|l|l|l|}
    \hline
        Class & Iforest & OC-SVM & PRAE-$\ell_0$ & PRAE-$\ell_1$ \\ \hline
        0 & 92.34 & 93.34 & 93.38 & 93.84 \\ \hline
        1 & 99.16 & 98.2 & 98.21 & 98.24 \\ \hline
        2 & 69.73 & 75.72 & 81.12 & 81.18 \\ \hline
        3 & 78.02 & 81.27 & 83.42 & 83.51 \\ \hline
        4 & 86.66 & 88.21 & 88.43 & 88.39 \\ \hline
        5 & 73.49 & 74.79 & 83.94 & 83.96 \\ \hline
        6 & 87.75 & 92.17 & 93.62 & 93.71 \\ \hline
        7 & 90.49 & 90.79 & 91.77 & 91.81 \\ \hline
        8 & 82.87 & 82.47 & 83.31 & 83.34 \\ \hline
        9 & 86.99 & 91.46 & 93.24 & 93.33 \\ \hline
         T-shirt/top & 90.45 & 90.62 & 90.65 & 90.84 \\ \hline
        Trouser & 97.77 & 97.83 & 97.96 & 97.90 \\ \hline
        Pullover & 86.26 & 85.27 & 87.24 & 86.66 \\ \hline
        Dress & 93.7 & 94.56 & 95.28 & 95.14 \\ \hline
        Coat & 91.95 & 91.17 & 92.47 & 92.15 \\ \hline
        Sandal & 92.12 & 91.58 & 91.14 & 91.15 \\ \hline
        Shirt & 80.48 & 80.25 & 80.86 & 80.88 \\ \hline
        Sneaker & 98.14 & 98.24 & 98.23 & 98.26 \\ \hline
        Bag & 87.14 & 84.42  & 84.49 & 85.58 \\ \hline
        Ankle boot & 97.65 & 98.16 & 97.95 & 97.91 \\ \hline \hline
         \large{\bf Average} & 88.15 & 89.03 & {\bf 90.34} & \color{blue}{\bf 90.39} \\ \hline
    \end{tabular}
    \caption{Performance comparison with two leading baselines (Iforest and OC-SVM) on the MNIST-S and Fashion MNIST datasets. The top ten rows correspond to different classes in MNIST; each row indicates the class label used to define the normal samples. The bottom ten rows correspond to different classes in Fashion MNIST. }
    \label{table:mist_all}
\end{table}
\subsection{Fashion MNIST}
To evaluate the ability to identify outliers in Fashion MNIST, we mix $5000$ nominal instances from the randomly selected 'Coat' class with $300$ anomalies sampled from all other fashion items. Evaluation of the AUC of our method compared to all baselines appears in Table 3. 
Next, we repeat this experiment using other classes as the majority/normal samples. 
In Table \ref{table:mist_all} we present the AUC of PRAE and two leading baselines across all classes. 
As evident from our result, the proposed approach outperforms leading baselines across most classes of Fashion MNIST.

\section{Running Time and Platform Details}
All the experiments were run on a server with an Intel(R) Xeon(R) CPU E5-2620 v3 @ 2.40GHz CPU and one GeForce GTX 1080 GPU with 8GB of memory. Our method scales like a standard autoencoder, and every iteration requires ${\cal O} (M+N)$ updates, where $M$ is the number of parameters in the network, and $N$ are the additional per sample parameters that serve as the anomaly score. Since modern NN are typically overparametrized, we argue that the additional $N$ parameters do not limit the use of our method since typically $M>N$. Across all examples used in this paper, a single run of our method does not take more than several minutes for the low-dimensional datasets and hours for the high-dimensional ones.

\end{document}